\documentclass[11pt, a4paper, copyright, gr]{google}
\usepackage[numbers, compress]{natbib}
\usepackage{graphicx}
\usepackage{soul}
\usepackage{xspace}
\usepackage{enumitem}
\usepackage{tikz}
\usepackage{hyperref}
\usepackage{xcolor}
\usepackage{wrapfig}
\definecolor{googleBlue}{HTML}{4285F4}
\definecolor{googleRed}{HTML}{EA4335}
\definecolor{googleYellow}{HTML}{FBBC05}
\definecolor{googleGreen}{HTML}{34A853}

\newcommand{\googleline}{%
  \noindent
  \begin{tikzpicture}[x=\linewidth, y=1pt]
    \def\slant{-0.05}
    \fill[googleBlue] (0,0) -- (0.25+\slant,0) -- (0.25,1) -- (0,1) -- cycle;
    \fill[googleRed] (0.25+\slant,0) -- (0.5+\slant,0) -- (0.5,1) -- (0.25,1) -- cycle;
    \fill[googleYellow] (0.5+\slant,0) -- (0.75+\slant,0) -- (0.75,1) -- (0.5,1) -- cycle;
    \fill[googleGreen] (0.75+\slant,0) -- (1,0) -- (1,1) -- (0.75,1) -- cycle;
  \end{tikzpicture}\par
}
\hypersetup{
    colorlinks=true,       % true = colored text; false = colored boxes
    linkcolor=orange, % Internal links (sections, etc.)
    citecolor=teal, % Citations
    urlcolor=blue!80!black,  % URLs
    filecolor=magenta,     
}
\usepackage[utf8]{inputenc} % allow utf-8 input
\usepackage[T1]{fontenc}    % use 8-bit T1 fonts
\usepackage{url}            % simple URL typesetting
\usepackage{booktabs}       % professional-quality tables
\usepackage{amsfonts}       % blackboard math symbols
\usepackage{nicefrac}       % compact symbols for 1/2, etc.
\usepackage{microtype}      % microtypography
\usepackage{xcolor}         % colors
\usepackage{multirow}
\usepackage{multicol}
\usepackage{tabularx}
\usepackage{ragged2e}
\usepackage{makecell}
\usepackage{amsmath}
\usepackage{cleveref}
\usepackage{subcaption}
\usepackage{colortbl} % add to preamble
\crefname{table}{table}{tables}
\Crefname{table}{Table}{Tables}

\uselogo{} 

\renewcommand{\titlefont}{\color{black}\normalfont\bfseries\fontsize{20}{22}\selectfont}

\title{WavesFM: Hierarchical Representation Learning for Longitudinal Wearable Sensor Waveforms}

\correspondingauthor{\{zjyang, tennisonliu, paolodia, mingzher\}@google.com}

\author[1,2,*]{Peng Cao}
\author[1,*,$\dagger$]{Zhijian Yang}
\author[1,*,$\dagger$]{Tennison Liu}
\author[1]{Jonathan Wang}
\author[1]{Jiang Wu}
\author[1]{Magdalena Proszewska}
\author[1]{Arvind Pillai}
\author[1]{Mingwu Gao}
\author[1]{Amir Farjadian}
\author[1]{Lawrence Cai}
\author[1]{Emily Blanchard}
\author[1]{Daniel McDuff}
\author[1]{Pramod Rudrapatna}
\author[1]{Matthew Thompson}
\author[1]{Anupam Pathak}
\author[1]{Mark Malhotra}
\author[1,3]{Shwetak Patel}
\author[2]{Dina Katabi}
\author[1,$\circ$,$\dagger$]{Paolo Di Achille}
\author[1,$\circ$,$\dagger$]{Ming-Zher Poh}

\affil[*]{Equal contributions}
\affil[$\circ$]{Joint supervision}
\affil[$\dagger$]{Corresponding Author}
\affil[1]{Google Research}
\affil[2]{MIT}
\affil[3]{University of Washington}

\begin{abstract}
Wearable sensors enable the continuous acquisition of high-resolution physiological waveforms, such as photoplethysmography~(PPG) and accelerometry~(ACC), under free-living conditions. However, inferring health-related phenotypes from these signals presents significant challenges due to high sampling frequencies, multimodal dependencies, and extreme sequence lengths (e.g., weeks of recordings), compounded by a scarcity of ground-truth labels. To address these challenges, existing \textit{self-supervised learning} (SSL) methodologies typically follow two paradigms: (1) learning rich \textit{morphological} representations from short waveform segments while collapsing longitudinal dynamics through simple aggregation, or (2) modeling \textit{behavioral} patterns from coarse, hand-crafted features (e.g. heart rate, step counts) spanning longer horizons but foregoing subtle, predictive signatures in raw waveforms. To bridge this gap, we propose \texttt{WavesFM}, a foundation model utilizing a two-stage SSL framework for longitudinal physiological data. Specifically, we decompose the learning problem into two stages: first, a segment-level encoder is pretrained to extract local embeddings from short waveforms; subsequently, a temporal encoder is trained to model the sequence of these embeddings across a multi-day horizon. This hierarchical approach overcomes the computational complexity of high-resolution, long-sequence data, allowing the overall model to capture both local signal semantics and the complex circadian and inter-day variations governing physiological dynamics. Pretrained on over $6.8$M hours ($N$=$324$k individuals) of recordings for the first stage and $5.3$M hours ($N$=$10$k) for the second stage, \texttt{WavesFM} demonstrates superior performance across $58$ diverse tasks spanning demographics, lifestyle, health conditions, and medications. 
\end{abstract}

\begin{document}

\maketitle

\section{Introduction}

The digitization of human physiology via wearable biosensors (e.g., photoplethysmography [PPG] and accelerometry [ACC]) is shifting modern healthcare from episodic snapshots to continuous, free-living monitoring for weeks or months at a time~\citep{dunn2018wearables,topol2019high}. These high-resolution longitudinal data offer an unprecedented window into multi-scale physiological dynamics, capturing rhythms that span orders of magnitude—from millisecond-scale heartbeat morphology to 24-hour circadian and multi-day homeostatic rhythms~\cite{pittman2013role,crnko2019circadian,kario2010morning}. Clinically meaningful phenotypes are defined not by static values, but by \textit{dynamic trajectories} over intra- and multi-day horizons~\cite{halimeh2022wearable,hall2018glucotypes,berkebile2025wearable}.

Translating these trajectories into clinical insights is hindered by two bottlenecks. First, the \textit{computational receptive field} required to process multimodal high-resolution signals over long horizons is prohibitively large (e.g., a single 100 Hz sensor yields roughly $10^8$ data points per week). Second, a profound \textit{signal-label asymmetry} exists between dense data streams and scarce subject-level clinical annotations, making end-to-end supervised learning infeasible. While \textit{self-supervised learning} (SSL) partially mitigates label scarcity, effectively distilling high-resolution, long-sequence waveforms into discriminative embeddings remains an open challenge.

Significant progress has been made in developing foundation models for wearable biosignals via SSL. These methodologies typically diverge into two paradigms to handle high-dimensional longitudinal data, exposing a fundamental \textit{resolution trade-off}. On the one hand, \textit{morphology-centric} approaches extract rich representations from short waveform segments (e.g. seconds to minutes) \cite{abbaspourazad2024largescale,abbaspourazad2024wearable,pillai2025papagei,saha2025pulse,nie2025anyppg,chen2025gpt,lee2025himae,gu2026cardiac} but collapse long-term dynamics via simple pooling. This treats multi-day data as an unordered ``bag of segments'', discarding vital temporal dependencies. Conversely, \textit{behavioral-centric} approaches capture long horizons by operating on coarse, aggregated metrics like minute-level mean heart rate or hourly step counts~\cite{narayanswamy2025scaling,xu2025lsm,erturk2025beyond}. While computationally scalable, this abstraction sacrifices subtle, predictive physiological signatures embedded within raw waveforms. Consequently, joint modeling high-frequency morphology and long-term longitudinal structure remains an unresolved challenge.

Here, we introduce \texttt{WavesFM}, a foundation model that reconciles this trade-off by decoupling local morphological encoding from long-horizon temporal modeling within a hierarchical SSL framework. Stage I comprises a \textit{segment encoder} that learns local, multimodal representations of high-frequency PPG and ACC segments via subject-contrastive learning. In Stage II, a \textit{temporal encoder} is trained to model the temporal dynamics of these segment-level embeddings over multi-day horizons using a masked reconstruction objective. To prevent the model from exploiting shortcuts inherent in latent-space masked modeling, we architecturally introduce a dual-branch decoder and multi-scale sparse masking. \texttt{WavesFM} yields a unified, subject-level embedding that integrates micro-scale morphology with macro-scale rhythms, bypassing the need to sacrifice resolution for context.

\begin{figure*}[t!]
	\centering
	\includegraphics[width=\linewidth, keepaspectratio]{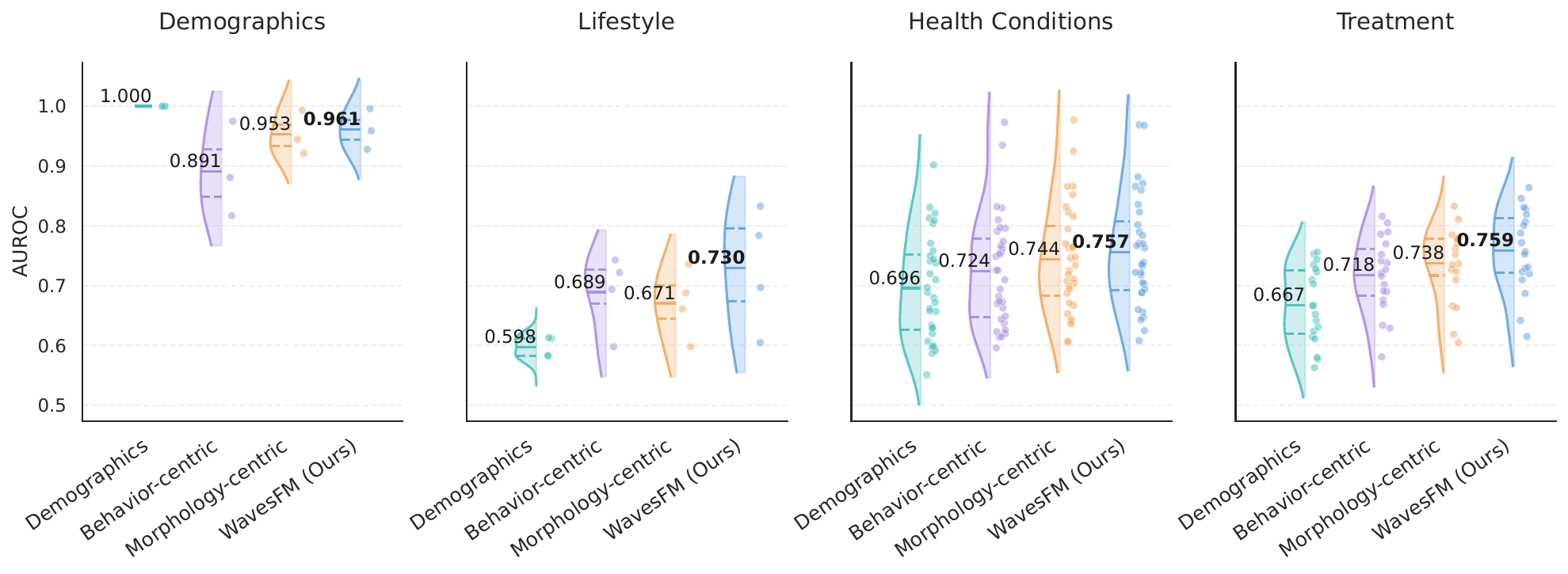}
	\caption{\textbf{Downstream performance by category.}~Compared to behavior-centric and morphology-centric approaches, \texttt{WavesFM} achieves superior performance across diverse downstream task categories by jointly modeling local morphology and longitudinal structure via two-stage hierarchical SSL.}
	\label{fig:teaser}
    \vspace{-0.5em}
    \googleline
    \vspace{-0.5em}
\end{figure*}

\textbf{Contributions.}~As such, the main contributions of this work are as follows:
\vspace{-1em}
\begin{enumerate}[leftmargin=*]
    \item \textbf{Technically}, we introduce \texttt{WavesFM}, a hierarchical foundation model decoupling local morphological encoding from long-horizon temporal modeling, trained via a two-stage SSL framework that combines morphological subject-contrastive objectives with masked embedding prediction.
    \item \textbf{Conceptually}, we demonstrate that resolving the trade-off between micro-scale waveform morphology and macro-scale temporal rhythms leads to more accurate capture of health phenotypes.
    \item \textbf{Empirically}, through large-scale validation across $58$ downstream tasks spanning demographics, health conditions, and medications, we demonstrate that \texttt{WavesFM} consistently outperforms existing models alongside morphological and behavioral approaches. We show the learned representations effectively capture physiologically grounded temporal variations and are robust to temporal missingness, crucial for health phenotyping in continuous wearable monitoring.
\end{enumerate}
\newcolumntype{L}{>{\RaggedRight\arraybackslash}X}
% Setup makecell to use bold font for headers by default
\renewcommand\theadfont{\bfseries}

\begin{table}[t!]
\centering
\footnotesize 
\caption{\textbf{Comparison of \texttt{WavesFM} with existing paradigms in wearable sensor representation learning.}}
\vspace{-0.5em}
\label{tab:paradigm_comparison}
\begin{tabularx}{\textwidth}{@{} 
    p{2cm} 
    p{2.4cm} 
    p{3.4cm} 
    p{1.8cm} 
    p{2.4cm} 
    p{2.4cm} 
@{}}
\toprule
\thead[l]{Paradigm} & \thead[l]{Key \\References} & \thead[l]{Input Fidelity \\ (Resolution)} & \thead[l]{Local \\Modeling} & \thead[l]{Temporal \\ Context} & \thead[l]{Temporal \\Modeling} \\ \midrule
Morphology-Centric & \cite{abbaspourazad2024largescale,abbaspourazad2024wearable,pillai2025papagei,saha2025pulse,nie2025anyppg,chen2025gpt,lee2025himae,gu2026cardiac} & \textbf{High: Raw Waveforms ($\mathbf{64}$-$\mathbf{100}$ Hz)} & \textbf{Self-supervised} & Short: $10$-$60$ sec & Mean\,Pooling (Stateless) \\ \midrule

Behavior-Centric & \cite{narayanswamy2025scaling,xu2025lsm,erturk2025beyond} & Low: Minute-hourly features (e.g., HR, steps) & Handcrafted features & Medium: $5$-$24$ hr \newline Long: 1 week & \textbf{Self-supervised (Stateful)} \\ \midrule

\textbf{Hierarchical SSL} & \textbf{\texttt{WavesFM} (Ours)} & \textbf{High: Raw Waveforms ($\mathbf{100}$ Hz)} & \textbf{Self-supervised} & \textbf{Long: $\mathbf{1}$ week} & \textbf{Self-supervised (Stateful)} \\ \bottomrule
\end{tabularx}
\end{table}

\section{Preliminaries}
\subsection{Problem Definition}
\label{subsec:problem_definition}
\textbf{Setup.}~Let $\mathbf{x} \in \mathbb{R}^{C\times L}$ represent a multi-day physiological recording, where $C$ denotes the number of concurrent sensor channels (e.g., PPG and ACC). The total sequence length $L$ is defined by the recording duration $T$ (in seconds) and sampling frequency $f_s$, such that $L = T \times f_s$. We assume access to a large-scale unlabeled dataset $\mathcal{D}_u = \{\mathbf{x}_j\}_{j=1}^N$ containing $N$ distinct temporal sequences, collected across subjects and disjoint time intervals. Additionally, we consider a smaller labeled dataset $\mathcal{D}_l = \{(\mathbf{x}_j, y_j)\}_{j=1}^M$, where $y_j$ denotes a subject-level health phenotype (e.g., a clinical diagnosis). Reflecting the real-world scarcity of health annotations, we assume $M \ll N$.

\textbf{Temporal decomposition.}~To handle the multi-scale nature of physiological data, we partition the global sequence $\mathbf{x}$ of length $L$ into a sequence of $K$ non-overlapping \textit{segments} $\mathcal{S}=\{\mathbf{s}_1,\mathbf{s}_2,\ldots,\mathbf{s}_K\}$, where each segment $\mathbf{s}_k \in\mathbb{R}^{C\times l}$ captures a short window of duration $\tau \ll T$ (e.g., $15$ seconds). Here $l=\tau \times f_s$ is the local segment length and $K = L/l$ is the total number of segments.

\textbf{Learning problem.}~Our objective is to learn a parameterized model $f_\theta: \mathbb{R}^{C\times L} \rightarrow \mathbb{R}^d$ using $\mathcal{D}_u$ that maps the high-dimensional raw signal $\mathbf{x}$ to a fixed $d$-dimensional representation $\mathbf{z} \in \mathbb{R}^d$. This representation ideally captures both local morphological signatures and global longitudinal structure over the horizon $T$. For a given downstream task, we learn a task-specific head $h_\xi:\mathbb{R}^d\rightarrow\mathcal{Y}$ that maps the frozen embedding $\mathbf{z}$ to the label space $\mathcal{Y}$ using the labeled dataset $\mathcal{D}_l$.

\textbf{Learning challenges.}~The learning problem is constrained by two primary factors:
\vspace{-0.5em}
\begin{enumerate}[leftmargin=*, itemsep=3pt, topsep=3pt, parsep=0pt]
    \item \textbf{Computational complexity.}~The input complexity of longitudinal sequences is $\mathcal{O}(C\cdot L)$. For reference, a week-long recording at $f_s = 100$Hz, $L=604,800$ seconds $\times$ $100$Hz produces $\approx 6\times10^7$ samples. Naive end-to-end processing of such sequences is computationally intractable.
    \item \textbf{Label scarcity.}~The imbalance between input dimensionality $C\times L$ and the number of labels $M$ precludes effective supervised learning, necessitating self-supervised approach to learn meaningful physiological and behavioral priors.
\end{enumerate}
\subsection{Wearable Biosignals}
Driven by their low power consumption, consumer wearables have largely converged on two primary biosensor modalities~\cite{kim2023photoplethysmography,yang2010review}: photoplethysmography (PPG) to measure blood volume changes, and tri-axial accelerometry (ACC) to track kinematic motions and posture. These modalities are frequently paired, as ACC provides a reference to isolate motion artifacts~\cite{lee2020motion}, and their combination captures the coupling between physical exertion and cardiovascular response~\cite{tonello2016correlates,sokas2023wearable}.

\textbf{Timescales.}~We model these signals jointly across two distinct scales, $\tau$ and $L$, aligned with underlying biological and behavioral rhythms. We set the local segment window $\tau$ to $15$ seconds, an epoch duration sufficient to encompass multiple cardiac cycles and resolve heartbeat morphology~\cite{sturge2026development}. This window is also sufficient to characterize stationary behavioral states and consistent kinematic patterns, such as a consistent gait or posture~\cite{liu2021assessment}.
We define the global temporal horizon $L$ as one week. 
While a $24$-hour window captures circadian oscillations (e.g., sleep-wake cycles), a seven-day horizon observes circaseptan patterns~\cite{de2026multiday}, including variations between workdays and weekends and crucial to disambiguating transient lifestyle fluctuations from invariant health phenotypes~\cite{vsimaityte2019objective,brooks2021impact}. 

\subsection{Related Works}
 
Recent years have witnessed substantial efforts toward self-supervised foundation models for wearable biosignals. These methods vary fundamentally in how they manage the extreme sequence lengths $L$ inherent in physiological data. 
This divergence has created a \textit{resolution trade-off}, where existing works prioritizes either high-frequency morphological resolution or longitudinal behavioral context, but rarely both. We summarize key distinctions to prior works in \Cref{tab:paradigm_comparison}.

\textbf{Morphology-centric} models focus on learning representations from short segments $\mathbf{s}_k$ (typically, $\tau \in [10\text{s}, 5\text{m}]$). For approaches that only consider PPG modeling, research has shown that self-supervised pretraining on large-scale data can encode rich cardiovascular, metabolic, and demographic information \cite{abbaspourazad2024largescale,saha2025pulse}. One predominant paradigm is contrastive learning, where distinctions are drawn regarding how positive pairs are constructed, including quality-invariance \cite{ding2024siamquality}, morphology-aware contrast \cite{pillai2025papagei}, and cross-model supervision from paired ECG \cite{nie2025anyppg}. Beyond contrastive objectives, alternative approaches have explored generative autoregressive training \cite{chen2025gpt}, masked reconstruction \cite{lee2025himae}, and multimodal masking \cite{gu2026cardiac}. In the domain of accelerometry modeling, approaches employ pretext task prediction \cite{yuan2024self}, relative contrastive learning \cite{xu2025relcon}, and knowledge distillation \cite{abbaspourazad2024wearable}. To our knowledge, no existing foundation model leverages joint modeling of PPG and ACC waveforms. More crucially, while these segment-level approaches capture fine-grained morphological detail, longitudinal structure is recovered only post-hoc via simple aggregation (e.g., mean-pooling \cite{abbaspourazad2024largescale, wu2026wearable}), which collapses the multi-day variations that are themselves clinically informative.

\textbf{Behavior-centric} models occupy the opposite extreme of the tradeoff, navigating the complexity of $L$ through dimensionality reduction. Rather than modeling the raw segments $\mathbf{s}_k\in\mathbb{R}^{C\times l}$, they operate on sequences of coarse, hand-crafted features $\tilde{\mathbf{s}}_k \in \mathbb{R}^p$ (e.g., minute-level heart rate, or step-count statistics), where $p \ll C\times l$. Notable approaches include \citet{narayanswamy2025scaling} and \citet{xu2025lsm}, which model sequences of these features across hours to a day, with the latter explicitly addressing data missingness. Other efforts have extended this line of research by aligning one-day feature matrices with natural-language captions \cite{zhangsensorlm}, or pursuing week-long behavioral streams \cite{erturk2025beyond}. While these models successfully capture multi-day rhythms, they forego predictive morphological signatures (e.g., pulse morphology) that can only be resolved from raw waveforms. 
\begin{figure}[t]
    \centering
    \begin{subfigure}{\linewidth}
        \centering
        \includegraphics[width=\linewidth]{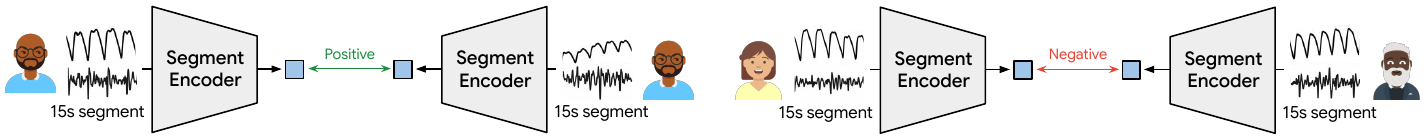}
        \caption{\textbf{Stage I: Local Representation Learning}}
        \label{fig:method_stage1}
    \end{subfigure} \\
    \vspace{1em}
    \begin{subfigure}{\linewidth}
        \centering
        \includegraphics[width=\linewidth]{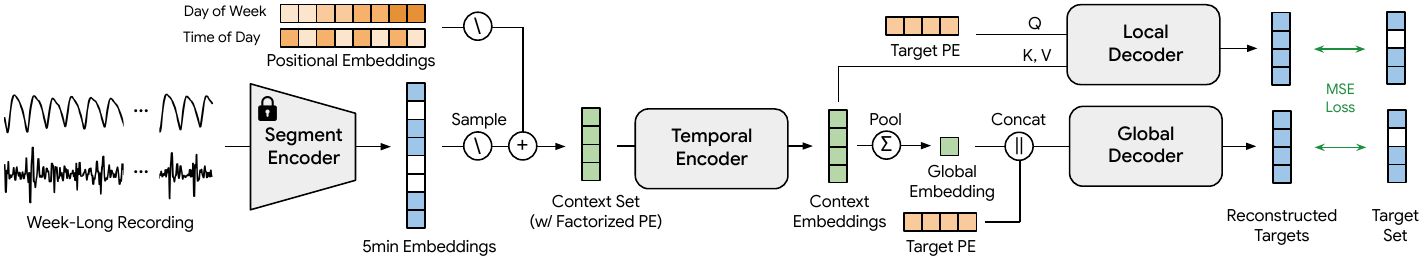}
        \caption{\textbf{Stage II: Temporal Dynamics Learning}}
        \label{fig:method_stage2}
    \end{subfigure}
    \caption{\textbf{Overview of the WavesFM framework.} (a)~Stage I utilizes a \textit{segment encoder} to learn \textit{segment embeddings} from $15$-s PPG and ACC windows via subject-contrastive learning, pairing intra-subject segments against those from different individuals. (b)~Stage II models long-horizon dynamics by pooling a week-long sequence of these segment embeddings into $5$-minute bins, augmenting them with circadian/circaseptan positional embeddings, and passing them to a \textit{temporal encoder} to output \textit{temporal embeddings}. During pretraining, dual-decoder branches reconstruct held-out targets: a local branch uses cross-attention with target positional encodings, while a global branch aggregates context through a mean-pooled bottleneck. The final week-level representations are utilized for downstream health phenotyping.} 
    \label{fig:method}
    \vspace{-0.5em}
    \googleline
    \vspace{-0.5em}
\end{figure}
\section{Method}

\textbf{Overview.}~We propose \texttt{WavesFM}, a hierarchical foundation model designed to learn representations from week-long continuous wearable waveform data. Our architecture rests on two core principles. First, we decouple morphological encoding from temporal modeling. As physiological states are quasi-stationary at the second scale but highly dynamic over multi-day horizons, this separation allows each encoder to specialize while drastically reducing the computational burden of high-resolution processing. Accordingly, the Stage I segment encoder (\Cref{subsec:method_stage_i}) maps $15$s multimodal waveforms into a denoised embedding space using a subject-contrastive objective~\cite{abbaspourazad2024largescale}, filtering out transient artifacts in favor of robust morphological features. Second, we model long-horizon dynamics directly within this latent space. The Stage II temporal encoder (\Cref{subsec:method_stage_ii}) models week-long sequence of segment embeddings via masked reconstruction, focusing on high-level semantic transitions rather than fine-grained signal details. To encourage meaningful temporal modeling, we employ multi-scale temporal masking alongside a dual-branch decoder. \Cref{fig:method} visualizes the overall framework.

\subsection{Stage I: Representation Learning on Multimodal Waveform Segment}
\label{subsec:method_stage_i}

The objective of Stage I is to map high-resolution multimodal segments $\mathbf{s}_k \in \mathbb{R}^{C\times l}$ into a $d$-dimensional latent space that preserves morphological features while suppressing stochastic noise. By training on short $\tau=15$s windows, we ensure that segment encoder $e_\phi$ captures locally stationary features (e.g., dicrotic notch in PPG or gait rhythm in ACC).

\textbf{Subject-contrastive objective.}~Standard contrastive frameworks often rely on aggressive data augmentation (e.g., jittering, scaling) to create positive pairs. While effective for images with strong spatial correlation, these transformations can inadvertently destroy subtle physiological signatures in physiological waveforms. 
To circumvent this, we define positive pairs as temporally disjoint segments sampled from the same subject, while negative pairs are segments sampled from different subjects. 
This formulation forces the model to learn to differentiate subject-specific physiological identity. 
Formally, let $\texttt{id}(\mathbf{s})$ denote the unique subject identity from which the segment $\mathbf{s}$ was sampled. A pair $(\mathbf{s}_i, \mathbf{s}_j)$ is then a positive pair if $\texttt{id}(\mathbf{s}_i) = \texttt{id}(\mathbf{s}_j)$ and $i\neq j$~\cite{abbaspourazad2024largescale}.

The segment encoder $e_\phi: \mathbb{R}^{C\times l} \rightarrow \mathbb{R}^d$ maps a segment $\mathbf{s}_k$ to an embedding $\mathbf{e}_k = e_\phi(\mathbf{s}_k)$. This embedding is projected to a contrastive manifold via a projection head $g_\omega: \mathbb{R}^d \rightarrow \mathbb{R}^{d'}$, such that $\mathbf{q}_k=g_\omega(\mathbf{e}_k)$, where $\mathbf{q}_k$ is also $\ell_2$-normalized. Concretely, for a batch of $B$ segments, let $\mathcal{I}=[B]$ be the set of indices. For an anchor segment $i \in \mathcal{I}$, we denote the set of positive indices $\mathcal{P}(i) = \{j\in\mathcal{I} \setminus \{i\} \::\:\texttt{id}(\mathbf{s}_i) = \texttt{id}(\mathbf{s}_j)\}$. The contrastive loss for anchor $i$ is then~\cite{oord2018representation,chen2020simple}:
\begin{equation}
\mathcal{L}_{i}^{\text{Stage I}} = -\frac{1}{|\mathcal{P}(i)|} \sum_{j \in \mathcal{P}(i)} \log \frac{\exp(\mathbf{q}_i \cdot \mathbf{q}_j / \gamma)}{\sum_{a \in \mathcal{I} \setminus \{i\}} \exp(\mathbf{q}_i \cdot \mathbf{q}_a / \gamma)},
\end{equation}
where $\gamma$ is the temperature hyperparameter. This objective forces the latent space to cluster by invariant physiological traits, effectively performing a semantic denoising step where only stable and robust morphological features are preserved. After Stage I pretraining, the projection head $g_\omega$ is discarded, and the encoder $e_\phi$ is frozen to produces segment embeddings for Stage II.

\subsection{Stage II: Temporal Dynamics Learning via Masked Embedding Prediction}
\label{subsec:method_stage_ii}

Stage II trains a temporal encoder $e_\psi$ to model the longitudinal evolution of physiological states over a one-week horizon. While Stage I embeddings capture local morphology, the temporal encoder captures macro-scale rhythms including circadian oscillations, sleep-wake transitions, and circaseptan (weekly) behavioral shifts. Specifically, $e_\psi: \mathbf{V}\mapsto\mathbf{z}$ maps a sequence of $M$ latent states $\mathbf{V}=[\mathbf{v}_1,\ldots,\mathbf{v}_M]^T \in \mathbb{R}^{M\times d}$ to a sequence-level representation $\mathbf{z} \in \mathbb{R}^d$.

\textit{Challenges of masked modeling on embedding sequences.}~Directly applying a masked-prediction objective to embedding sequences risks two failure modes that bypass meaningful temporal modeling: (i) \emph{Interpolation shortcut}: because slow-varying physiological states exhibit high temporal autocorrelation, missing embeddings can be trivially reconstructed from visible neighbors; (ii) \textit{Identity shortcut}: the encoder can degenerate into an identity function, passing context embeddings straight through to the decoder to perform heavy lifting and resolve target reconstruction.
We break the interpolation shortcut at the input stage using an aggressive, multi-scale masking strategy that creates contiguous missing patches. Simultaneously, we eliminate the identity shortcut by using dual-branch decoders, forcing the encoder to route information through a collapsed temporal bottleneck.

\textbf{Input sequence construction.}~To mitigate short-term redundancy and ensure computational tractability, we aggregate the $K$ segment embeddings from Stage I into a compressed sequence of $M = 2016$ temporal bins (representing one week at $5$-minute intervals). Each bin embedding $\mathbf{v}_t$ is computed as the mean of $20$ contiguous, non-overlapping Stage I embeddings. For time bins where data is missing—due to off-body periods or sensor duty-cycling—we replace $\mathbf{v}_t$ with a learnable $\texttt{[MISSING]}$ token. Conventional absolute positional encoding schemes are ill-suited for free-living data because recordings are sampled \textit{arbitrarily} relative to the calendar; a sequence may begin at any day or time, making absolute indices semantically inconsistent across samples. To explicitly encode the nested periodicities of human physiology, we introduce a \textit{factorized positional encoding} scheme anchored to the recording's calendar time, allowing the model to understand biologically recurring patterns. 

Let $w_t = \texttt{cal}(\mathbf{v}_t)$ be the calendar time of each embedding. We map $w_t$ to factorized indices via modulo functions $\sigma_{\text{dow}}(w_t) \in \{1, \dots, 7\}$ and $\sigma_{\text{tod}}(w_t) \in \{1, \dots, 288\}$, utilizing two learnable lookup tables, $\mathbf{P}_{\text{dow}} \in \mathbb{R}^{7 \times d}$ and $\mathbf{P}_{\text{tod}} \in \mathbb{R}^{288 \times d}$. These represent \textit{day-of-week} and \textit{time-of-day} information ($7$ days in a week; $288$ $5$-minute bins in a $24$-hour day), thus yielding:
\begin{equation}
    \mathbf{v}_t \leftarrow \mathbf{v}_t + (\mathbf{P}_{\text{dow}})_{\sigma_{\text{dow}}(w_t),:} + (\mathbf{P}_{\text{tod}})_{\sigma_{\text{tod}}(w_t),:}
\end{equation}
This factorization ensures that embeddings from identical calendar periods share the same positional encoding regardless of sequence index, injecting circadian/circaseptan priors into the temporal model.

\textbf{Multi-scale masking.}~To break the interpolation shortcuts inherent in autocorrelated longitudinal sequences, we employ an aggressive masking strategy. From the full sequence of $M$ embeddings, we sample a sparse \textit{context set} $\mathbf{C}$ of size $N_\text{ctx}=252$ (approximately $12.5\%$ of the weekly horizon) to serve as input to the temporal encoder $e_\psi$. The sampling of $\mathbf{C}$ follows a \textit{multi-scale patching protocol} designed to capture dynamics across different temporal reslutions. For each training sample, we randomly draw a patch size $P \sim \texttt{Uniform}(\{1, 2, 4\})$, corresponding to random patches of $5, 20, 60$-minutes, respectively. We then partition the index set $\{1,\ldots,M\}$ into $M/P$ non-overlapping contiguous patches; and randomly sample $N_\text{ctx}/P$ patches to form the context set. The target set is then $\mathbf{T}=\mathbf{V}\setminus\mathbf{C}$. This multi-scale strategy serves dual purposes: small patches forces the model to resolve fine-grained physiological transitions, such as sleep-stage shifts or instantaneous activity changes, while larger patches create extended missingness that demand multi-hour context integration, removing the dense-neighbor structure that interpolation relies on.

\textbf{Dual-branch reconstruction.} The context sequence, augmented with factorized positional encodings, is processed by the temporal encoder—a stack of Transformer blocks \cite{vaswani2017attention}—to produce sequence-level context embeddings. To eliminate the interpolation shortcut inherent in highly redundant time-series data, we decode these embeddings through two complementary branches with distinct geometric constraints: (i) a \textit{local} branch: a standard cross-attention decoder where target positional embeddings query the context embeddings to reconstruct each target position; (ii) a \textit{global} branch: the context embeddings are mean-pooled along the temporal dimension into a single $d$-dimensional week-level vector. This global descriptor is then concatenated with the target positional encodings and processed by a Transformer decoder to reconstruct the missing targets.

By collapsing the temporal sequence into a single bottleneck vector, the global decoder branch cannot perform localized lookups from adjacent unmasked bins, requiring the temporal encoder successfully compresses week-long health dynamics into a summary vector. Conversely, as relying solely on this bottleneck would over-smooth the latent space, the local decoder branch provides dense, per-token gradients for the temporal encoder to preserve fine-grained temporal structures.

\textbf{Reconstruction objective.}~We minimize the MSE between the ground-truth and predicted target embeddings.  Let $\mathcal{M} \subset [M]$ be the set of indices for reconstruction targets in $\mathbf{T}$, the combined loss is:
\begin{equation}
    \mathcal{L}^{\text{Stage II}} = \underbrace{\frac{1}{d\cdot |\mathcal{M}|}\sum_{\mathbf{t \in \mathcal{M}}} \left\|\mathbf{v}_t - \hat{\mathbf{v}}_t^{\text{l}}\right\|_2^2}_{\text{Local branch loss}} + \underbrace{\frac{1}{d\cdot|\mathcal{M}|}\sum_{t\in\mathcal{M}}\left\|\mathbf{v}_t - \hat{\mathbf{v}}_t^{\text{g}}\right\|^2_2}_{\text{Global branch loss}}
\end{equation}
where $\hat{\mathbf{v}}_t^{\text{l}}, \hat{\mathbf{v}}_t^{\text{g}}$ are the corresponding reconstructions from the local and global branches respectively.

\textbf{Downstream use.}~At inference time, a week-long signal is processed end-to-end: $15$s segments are encoded by the segment encoder, mean-pooled into $5$-minute bins, augmented with factorized positional embeddings, and passed through the temporal encoder. The resulting week-level embedding produced by the global decoder branch serves as final representations for downstream probing.
\begin{figure*}[t!]
	\centering
	\includegraphics[width=\linewidth, keepaspectratio]{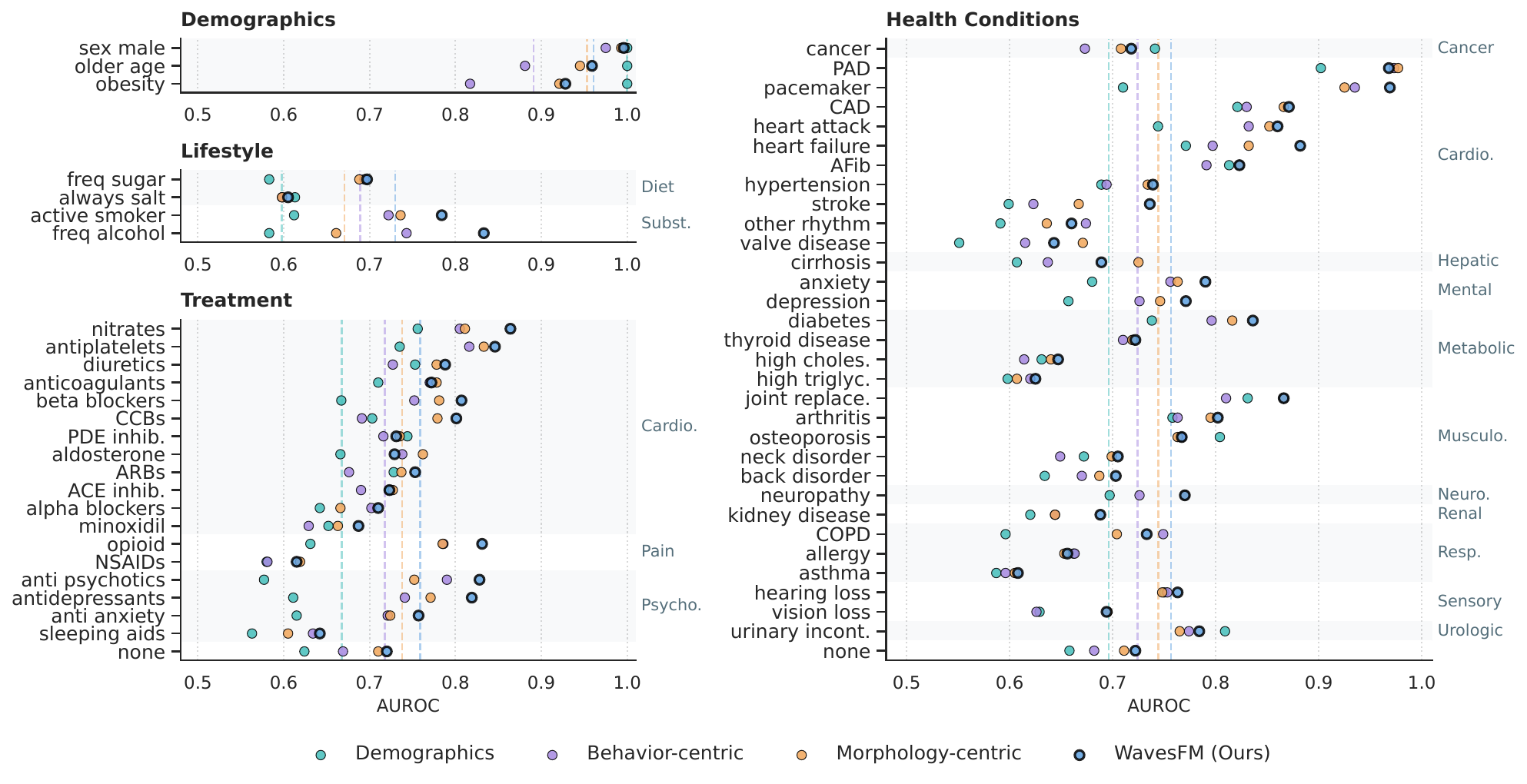}
    \vspace{-2em}
	\caption{\textbf{WavesFM outperforms baselines across clinical prediction tasks.} Test-set AUROC across demographics, lifestyle, treatment and health condition tasks are reported. Tasks are grouped by clinical categories~(right-side labels); dashed lines mark per-panel mean AUROC for each model.
    }
	\label{fig:main_results}
    \vspace{-0.5em}
    \googleline
    \vspace{-0.5em}
\end{figure*}

\section{Experiments}
\textbf{Pretraining and evaluation datasets.}~We implemented \texttt{WavesFM} and trained it on an extensive de-identified, unlabeled dataset of PPG and ACC waveforms. All participants provided voluntary consent for the use of their de-identified data, and a secondary research exemption was obtained from an Institutional Review Board (IRB). We used $6.8$M hours of intermittently sampled waveforms from $324$k individuals to train Stage I, and a subset of this data, $5.3$M hours from $61.2$k week-long recordings ($N$=$10$k individuals), to train Stage II. 
Downstream task probing and testing were conducted on a held-out dataset of $11,736$ participants, collected under informed consent from the Fitbit Hypertension Study using an IRB-approved study protocol, with an average of $11.7$ weeks of data per subject. Data was split at the subject level ($70/15/15$ for train/val/test) to prevent data leakage.
This evaluation set includes $58$ self-reported labels spanning demographics, lifestyle behaviors, health conditions, and treatments. More details on data splits and preprocessing are included in Appendix~\ref{sec:dataset_details}.

\textbf{Baselines.}~To evaluate the empirical performance of \texttt{WavesFM}, we compare it against two baselines representing the dominant paradigms in wearable representation learning, both trained on our dataset and constructed using components of our framework to isolate the contribution of hierarchical modeling from confounders such as pretraining data or model capacity: (i) a \emph{Morphology-Centric} baseline that mean-pools the frozen Stage I segment embeddings into a subject-level representation, preserving local signal fidelity but discarding multi-day temporal dynamics; and (ii) a \emph{Behavior-Centric} baseline that replaces our Stage I segment embeddings with handcrafted features derived using NeuroKit2~\cite{Makowski2021neurokit}~(e.g., heart rate, heart rate variability, motion metrics) and trains our Stage II temporal encoder on top, capturing multi-day behavioral shifts but forfeiting fine-grained morphological resolution. 
To establish a reference for predictive difficulty of each task, we additionally evaluate a demographics-only model using age, sex, and BMI as features. We also benchmark against supervised temporal aggregators trained directly on unpooled Stage-I embedding sequences and open-weight public PPG foundation models~\cite{pillai2025papagei,saha2025pulse,nie2025anyppg}. Baseline implementation details are reported in Appendix~\ref{sec:implementation_details}.

\textbf{Evaluation protocol.}~We formulate each of the $58$ labels as a binary classification task and evaluate all models under a standard linear probing protocol on top of frozen subject-level embeddings.
For the morphology-centric baseline, subject-level embeddings are obtained by mean-pooling all segment embeddings across each subject's recordings. 
For the behavior-centric baseline and \texttt{WavesFM}, week-level embeddings are extracted from the frozen temporal encoder for each available week and then averaged across all weeks per subject to yield a single subject-level representation.
For each task, we train an $\ell_2$-regularized logistic regression, tuning the regularization strength on validation set. 
Final performance is reported on the held-out test set.
We report AUROC as our primary metric in the main text; to assess statistical robustness, each metric is averaged over $1,000$ bootstrap resamples of the test set, with means reported in the main text and $95\%$ confidence intervals together with the partial AUROC (pAUC) at a $10\%$ false positive rate (FPR) threshold reported in Appendix~\ref{sec:detailed-results}.

\subsection{WavesFM Outperforms Morphology- and Behavior- Centric Baselines}

Figure~\ref{fig:main_results} shows downstream performance across all $58$ tasks. All physiology-based paradigms substantially exceed the demographics-only reference, confirming that wearable signals carry health-relevant information beyond age, sex, and BMI. 
\texttt{WavesFM} achieved the best average rank of $1.44$~(Figure~\ref{fig:stat_ranking}), outperforming all baseline models by a statistically significant margin under a post-hoc Nemenyi test ($\alpha$ = $0.05$, CD = $0.616$), and surpassing the behavior-centric baseline on $52$ of $58$ tasks ($89.7\%$) and the morphology-centric baseline on $46$ of $58$ tasks ($79.3\%$). 
Across categories (Figure~\ref{fig:teaser}), \texttt{WavesFM} yields mean AUROC improvements of $+7.0$ points on demographics, $+4.1$ on lifestyle, $+3.3$ on health conditions, and $+4.1$ on treatment over the behavior-centric baseline, with corresponding gains over the morphology-centric baseline of $+0.8$, $+5.9$, $+1.3$, and $+2.1$, respectively.

Notably, the largest gains over the behavior-centric baseline are found on tasks whose predictive signatures may live in waveform morphology: obesity ($0.817$$\to$$0.928$, $+13.6\%$), calcium channel blocker (CCB) use ($0.691$$\to$$0.801$, $+15.9\%$) and angiotension receptor blocker (ARB) use ($0.676$$\to$$0.753$, $+11.4\%$). Obesity is associated with increased arterial stiffness \cite{rider2010effect} whereas CCB and ARB reduce vascular resistance and pulse wave reflections \cite{safar2009central}, changing the PPG waveform morphology which cannot be captured by minute-level statistical features.
Conversely, the largest gains over the morphology-centric baseline concentrate on tasks whose signatures unfold over multi-day horizons rather than within individual segments, such as frequent alcohol use ($0.661$$\to$$0.833$, $+26.0\%$) and antipsychotics use ($0.752$$\to$$0.828$, $+10.1\%$).
These phenotypes may manifest as transient changes in the PPG waveform along with disruptions in circadian rhythms \cite{moon2021effects} or sleep architecture \cite{gardiner2025effect} that segment-level mean-pooling cannot resolve, but that the temporal encoder captures by integrating context across the full week-long horizon.
Together, these results demonstrate that jointly modeling micro-scale morphology and macro-scale temporal rhythms improves health phenotyping.

\begin{wrapfigure}{r}{0.49\textwidth}
    \vspace{-1.5em}
	\centering
	\includegraphics[width=\linewidth, keepaspectratio]{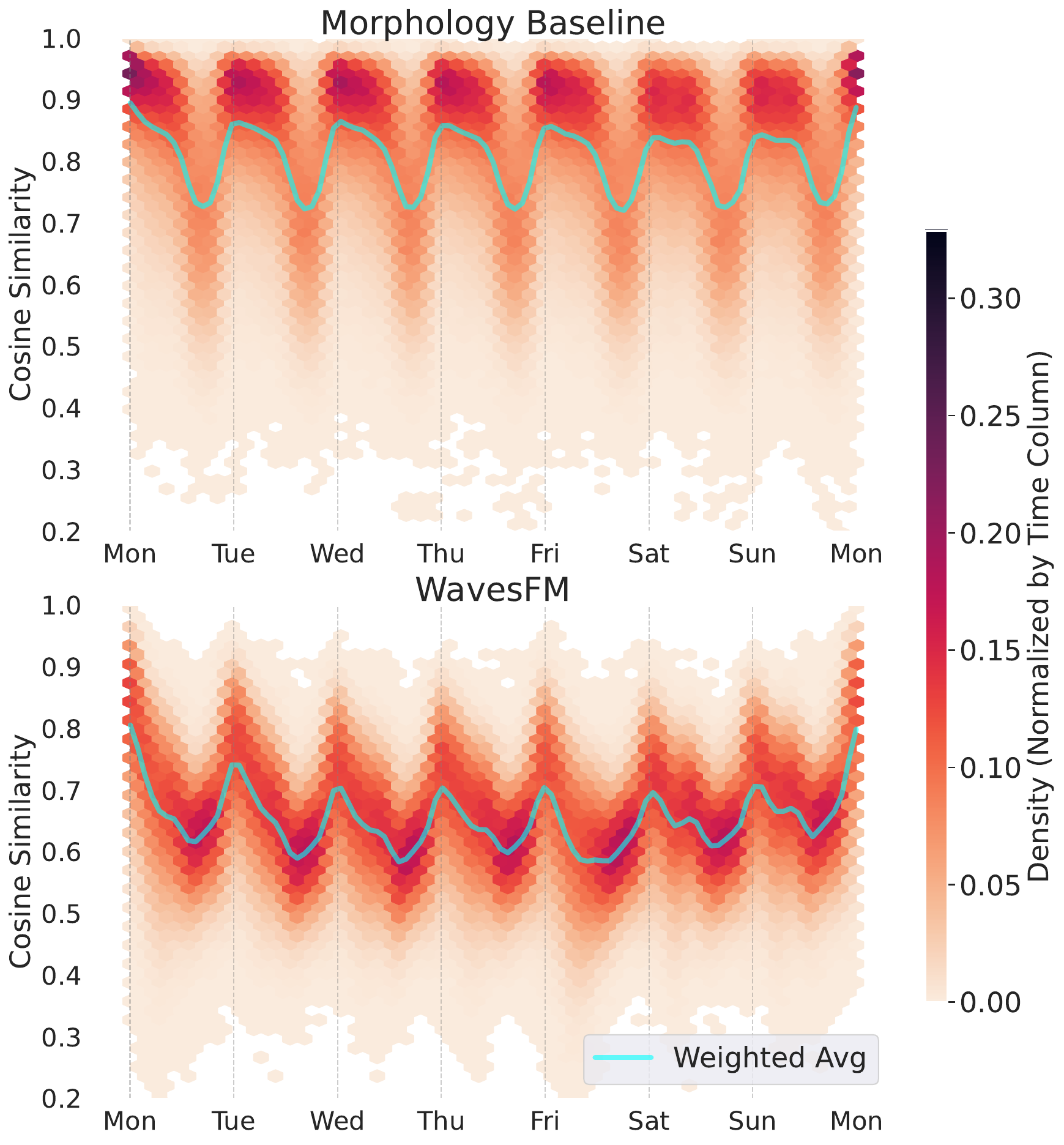}
	\caption{\textbf{Stage II embeddings exhibit stronger circadian structure.} Hexbin distributions of pairwise cosine similarities across the week, referenced to each subject's Monday $9$–$10$ AM bin.}
	\label{fig:circadian}
    \vspace{-0.5em}
\end{wrapfigure}
\textbf{Learned temporal dynamics.}~To understand the representations driving these gains, particularly against the stronger morphology-centric baseline, we analyzed the unpooled, week-long embedding sequences at a 5-minute resolution. Specifically, we compared the baseline's aggregated Stage I segment embeddings against the per-token outputs of \texttt{WavesFM}'s Stage II temporal encoder.
As shown in \Cref{fig:circadian}, the Stage I embeddings produces a shallow binary pattern: similarity remains tightly clustered throughout the daytime hours and dips during sleep, indicating that segment-level morphology alone resolves little beyond a sleep/wake contrast.
The Stage II embeddings, in contrast, exhibit a continuous, full-cycle circadian rhythm—similarity peaks sharply at the $9$–$10$ AM reference window each day and descends smoothly to nighttime, recovering the physiological transitions that unfold between waking, peak activity, and rest. The weekday-to-weekend shift is also visible as a slight flattening of the Saturday–Sunday peaks, consistent with altered routines on non-work days.
More analyses are included in Appendix~\ref{sec:latent-space-analyses}, showing that the factorized positional embeddings recover circadian and circaseptan structure (\Cref{fig:positionalembedding}) and that the rhythmic geometry persists under alternative reference anchors and across inter-segment time distances (\Cref{fig:circadian_combined}).
Together, these confirm the temporal encoder's ability to model temporal dynamics across multiple scales, explaining \texttt{WavesFM}'s superiority on the tasks characterized by temporal variations.

\subsection{Robustness to Longitudinal Missingness}
\begin{wrapfigure}{r}{0.35\textwidth}
    \vspace{-1.1em}
	\centering
	\includegraphics[width=0.9\linewidth, keepaspectratio]{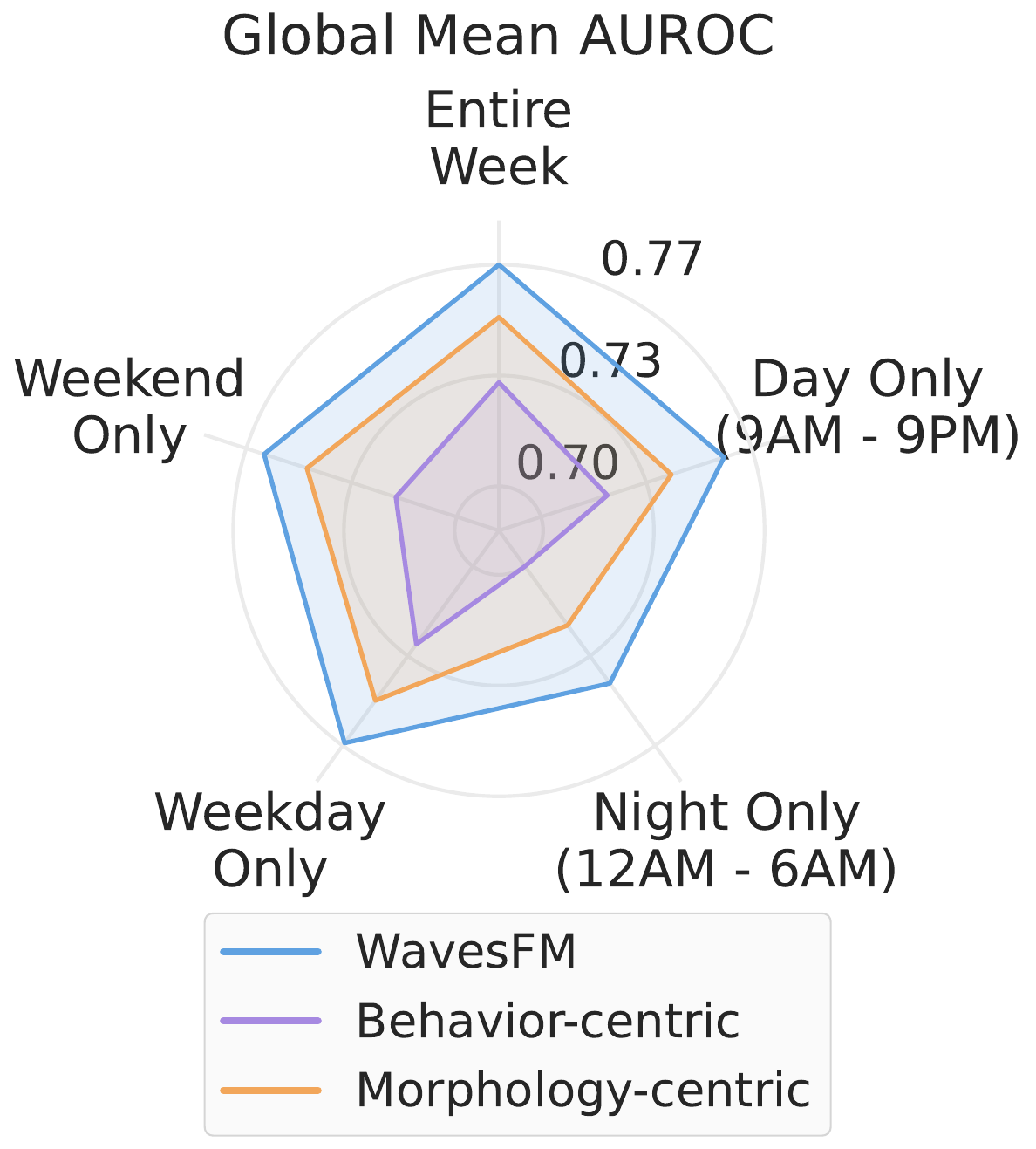}
    \vspace{-0.5em}
	\caption{\textbf{Performance across longitudinal missingness.}}
	\label{fig:ppgfm_radar_robustness_main}
    \vspace{-0.7em}
\end{wrapfigure}
Wearable recordings are routinely incomplete due to duty-cycling and irregular wear, so we evaluated each model under five systematic missingness scenarios: the entire week, daytime ($9$AM–$9$PM) only, nighttime ($12$AM–$6$AM) only, weekdays only, and weekends only. As shown in \Cref{fig:ppgfm_radar_robustness_main}, \texttt{WavesFM} achieves the highest mean AUROC across every scenario. The most challenging condition for all three models is night-only, where reduced physiological variance limits the discriminability of health-relevant signals; nonetheless, \texttt{WavesFM}'s AUROC drops by $3.3\%$ ($0.767$$\to$$0.742$) relative to the full-week baseline, compared to $4.1\%$ for the morphology-centric baseline ($0.750$$\to$$0.719$) and $4.7\%$ for the behavior-centric baseline ($0.729$$\to$$0.695$). A similar larger drop of $1.8\%$ is seen in the behavior-centric model during weekends ($0.729$$\to$$0.716$), potentially due to disrupted macro-behavioral routines, whereas \texttt{WavesFM} further demonstrates robustness with a drop of only $0.8\%$ ($0.767$$\to$$0.761$) by successfully capturing complementary high-resolution morphology. Per-category breakdowns for longitudinal missingness are included in Appendix~\ref{sec:missingness}.

\subsection{Hierarchical SSL Outperforms Supervised Baselines Under Label Scarcity}
\begin{wraptable}{r}{0.35\textwidth}
  \footnotesize
  \centering
  \vspace{-1.5em}
  \caption{\textbf{Comparison with supervised baselines trained on a $10\%$ subset.}}
  \begin{tabular}{lc}
    \toprule
    \textbf{Method} & \textbf{AUROC}$^\uparrow$ \\ %& \textbf{pAUC}$^\uparrow$ \\
    \midrule
    LSTM             & 0.658 \\ %& 0.567 \\
    Weighted Pooling & 0.661 \\ %& 0.561 \\
    Cross-attention  & 0.667 \\ %& 0.565 \\
    \midrule
    \texttt{WavesFM}        & \textbf{0.719} \\ %& \textbf{0.595} \\
    \bottomrule
  \end{tabular}
  \label{tab:supervised_baseline_overall}
\end{wraptable}
Since self-supervised temporal pretraining is motivated by the label-scarcity that characterizes wearable cohorts, we compare linear probing on \texttt{WavesFM}'s pretrained embeddings against supervised temporal aggregators (e.g., an LSTM, cross-attention, or weighted pooling) trained directly on the frozen Stage~I segment embeddings under label-scarce settings. 
We trained both the linear probe and each supervised aggregator on a $10\%$ subject-level subset of the labeled training data ($1$k subjects) and evaluate on the full held-out test set. 
As shown in \Cref{tab:supervised_baseline_overall}~(per-task results in \Cref{tab:supervised_baseline_detail}), none of the three aggregators matches the performance of a simple linear probe on \texttt{WavesFM}'s pretrained subject-level embeddings, with \texttt{WavesFM} achieving an average AUROC improvement of $5.2$ points over the strongest supervised baseline. By absorbing long-horizon dynamics into a single label-free representation during pretraining, the temporal encoder shifts the burden of long-sequence modeling away from the downstream task, enabling efficient linear probing in the label-scarce regimes.

\subsection{Comparison with Public PPG Foundation Models}

As an additional point of comparison, we evaluate \texttt{WavesFM} against three publicly available PPG foundation models: \texttt{PaPaGei} \cite{pillai2025papagei}, \texttt{Pulse-PPG} \cite{saha2025pulse}, and \texttt{AnyPPG} \cite{nie2025anyppg} in Appendix~\ref{sec:public-ppg-fm}. While \texttt{WavesFM} achieves stronger downstream performance, we caution against interpreting these results as a direct architectural comparison, as our evaluation cohort comes from a different data source than their pretraining data.
Robust transfer across sensor hardware and recording protocols remains an open problem for wearable foundation models.
Additionally, the segment encoder architectures underlying these models are largely complementary to our framework and could be substituted as the Stage I encoder, with our hierarchical Stage II providing the longitudinal modeling on top.

\subsection{Ablation Studies}
We empirically validate the architectural interventions introduced in Section~\ref{subsec:method_stage_ii} to prevent shortcut learning in Stage II in Appendix~\ref{sec:ablation}. 
Removing either branch of the dual-branch decoder degrades performance, with the gap widening under data constraints (night-only inputs, $10\%$ subject subsets), indicating that the two branches contribute complementary, non-redundant pressure on the encoder (\Cref{fig:ablation_dualbranch}). Furthermore, aggressive context subsampling (e,g, retaining only $12.5\%$ to $25\%$ of the sequence) yields significantly better representation than milder masking (\Cref{fig:ablation_context}), and utilizing an average-pooled segment embedding proves consistently superior to a learnable token for the week-level representation (\Cref{fig:ablation_cls}). Together, these confirm that our interventions are collectively necessary to close the interpolation and identity shortcuts.

\vspace{-1mm}
\section{Conclusion}

In summary, we introduced \texttt{WavesFM}, a foundation model for multimodal PPG and ACC biosignals that effectively addresses the trade-off between high-resolution signal morphology and longitudinal temporal context. By decoupling the learning process into a hierarchical two-stage framework, \texttt{WavesFM} utilizes a segment encoder to robustly extract morphological features and a temporal encoder to capture the multi-scale rhythms governing human physiology. Empirical validation across a diverse suite of clinical, demographic, and lifestyle tasks confirms that \texttt{WavesFM} significantly and consistently outperforms existing foundation models as well as specialized morphological and behavioral paradigms. Furthermore, the learned representations demonstrate robustness to temporal missingness, which is crucial for continuous wearable monitoring. More broadly, our analyses suggest that capturing the interaction between micro-scale morphological fidelity and macro-scale longitudinal dynamics is a vital direction towards more precise health phenotyping via consumer health wearables. 
Limitations, future work, and societal impacts are discussed in Appendix~\ref{sec:discussion}.

{
\small
\bibliographystyle{unsrtnat}
\bibliography{ref}
}
\clearpage
\appendix
\section{Discussion}
\label{sec:discussion}
\subsection{Limitations and Future Work}
\label{sec:limitation}
WavesFM has several limitations that point to future directions. First, the model assumes availability of both PPG and ACC channels from a single device type, and was not trained to accommodate missing modalities and may not generalize across heterogeneous hardware devices. 
Modality-dropout pretraining and cross-device adaptation are promising directions in future work to enable robust performance under missing channels and heterogeneous hardware.
Second, the subject-level design targets stable, persistent health conditions and has not been evaluated on tasks requiring finer temporal resolution, such as acute event detection or intra-day physiological dynamics. Finally, the dense subject-level embedding lacks clinical interpretability without offering an account of which signal characteristics drove a given prediction. Future work could address this limitation through post-hoc attribution or disentangled representation learning.

We are committed to the principles of open science and acknowledge the importance of open data in advancing scientific inquiry. We acknowledge that developing and validating our methods on closed, non-public datasets limits the scientific community from fully replicating our work. At the same time, given the sensitive nature of health data, these considerations must be balanced with protecting the privacy of our participants and the security of their personal health information. Although we understand that the inability to release this data constitutes a limitation, we are confident that the findings derived from it provide significant and beneficial insights to the research community.

\subsection{Broader Impacts}
\label{sec:broader-impacts}

WavesFM advances research on evaluating meaningful health information from continuous, passively collected wearable data, with several potential societal impacts. By enabling automated health characterization and condition monitoring from consumer-grade devices, the framework could potentially lower barriers to early disease detection and longitudinal health tracking, particularly for populations with limited access to frequent clinical care. Subject-level representations learned from raw waveforms may also accelerate clinical research by providing richer phenotyping tools for observational studies and clinical trials, reducing reliance on costly or invasive assessments.

\section{Dataset Details}
\label{sec:dataset_details}

In this section, we provide details regarding the pretraining and downstream datasets, as well as the associated signal processing pipeline. Across all datasets, we utilize de-identified PPG and ACC waveforms from participants who provided voluntary informed consent for the use of their data in the research and development of novel health and wellness features. Additionally, a secondary research exemption determination was obtained from an Institutional Review Board (IRB).

\begin{figure*}[h]
	\centering
	\includegraphics[width=0.75\linewidth, keepaspectratio]{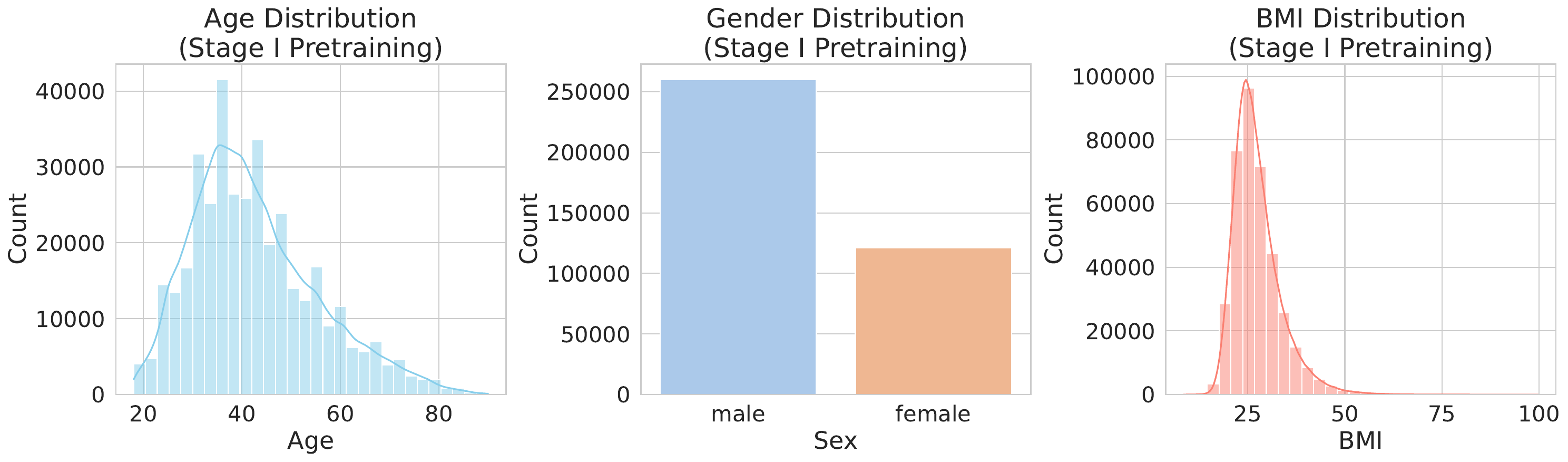}
	\caption{\textbf{Cohort demographics (Stage I pretraining dataset)}.}
	\label{fig:stage_1_demographics}
\end{figure*}

\textbf{Stage I pretraining dataset.}~For Stage I pretraining, the dataset was curated from $324$k individuals. For each participant, we intermittently sampled $5$k segments of $15$s duration, resulting in a corpus of approximately $6.8$M hours of recordings ($324$k individuals $\times$ $5$k segments $\times$ $15$s). Detailed demographic information for this cohort is provided in \Cref{fig:stage_1_demographics}.

\textbf{Signal processing.}~Raw $15$s signal segments are extracted for each subject. After computing the one-dimensional magnitude of the three-axis ACC data, both modalities are resampled to $f_s=100$Hz using a polyphase FIR filter (\texttt{scipy.signal.resample\_poly}). We then apply a standard FIR bandpass filter (\texttt{scipy.signal.firwin}), utilizing a passband of $1$-$12$Hz for the PPG signals and $0.5$-$49.0$Hz for the ACC magnitude. Finally, each channel is independently normalized using Z-score normalization.

\begin{figure*}[h]
	\centering
	\includegraphics[width=0.75\linewidth, keepaspectratio]{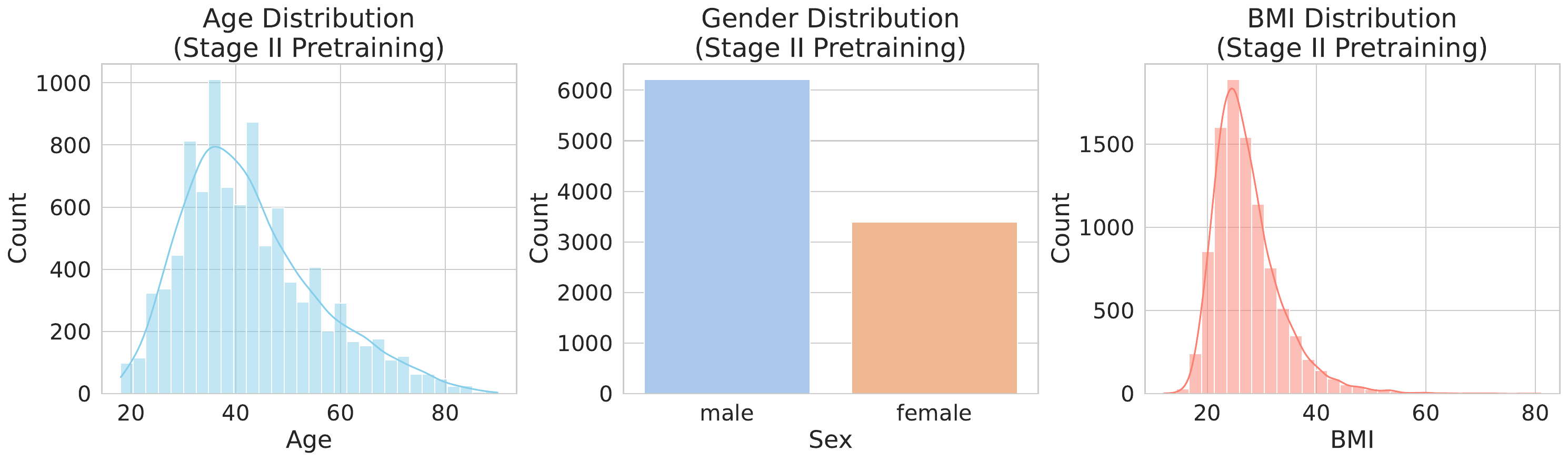}
	\caption{\textbf{Cohort demographics (Stage II pretraining dataset)}.}
	\label{fig:stage_2_demographics}
\end{figure*}

\textbf{Stage II pretraining dataset.}~The Stage II pretraining dataset consists of a longitudinal subset of the Stage I cohort. Specifically, we selected $10$k individuals with at least one week of available data, yielding $61.2$k unique week-long recordings and totaling $5.3$M hours (47.4\% missingness). Demographic information for this subset is detailed in \Cref{fig:stage_2_demographics}.

\begin{figure*}[h]
	\centering
	\includegraphics[width=\linewidth, keepaspectratio]{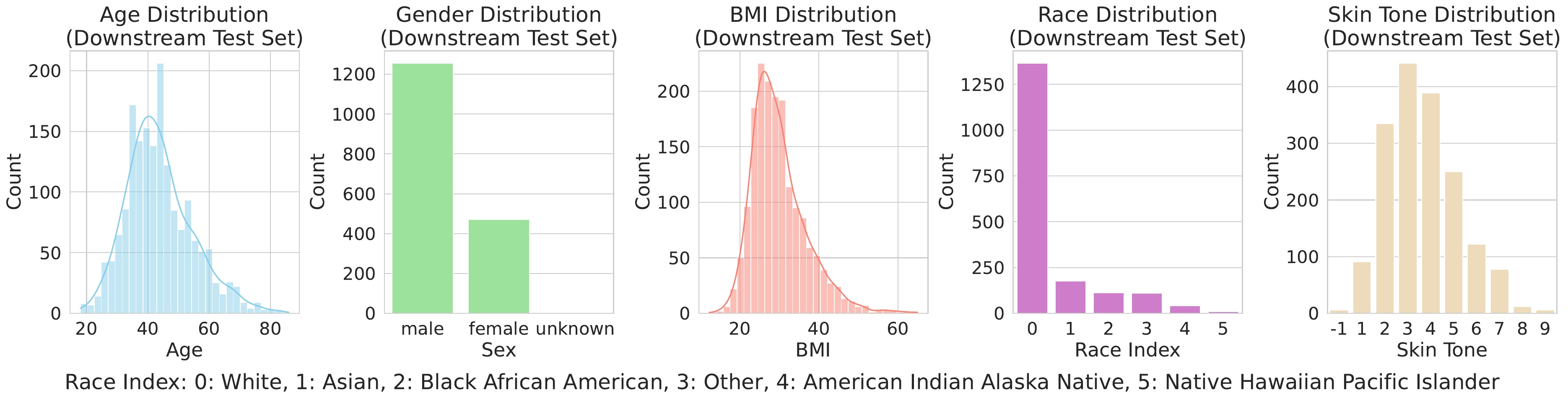}
	\caption{\textbf{Cohort demographics (downstream test split)}.}
	\label{fig:downstream_demographics}
\end{figure*}

\begin{figure*}[p]
	\centering
	\includegraphics[width=0.75\linewidth, keepaspectratio]{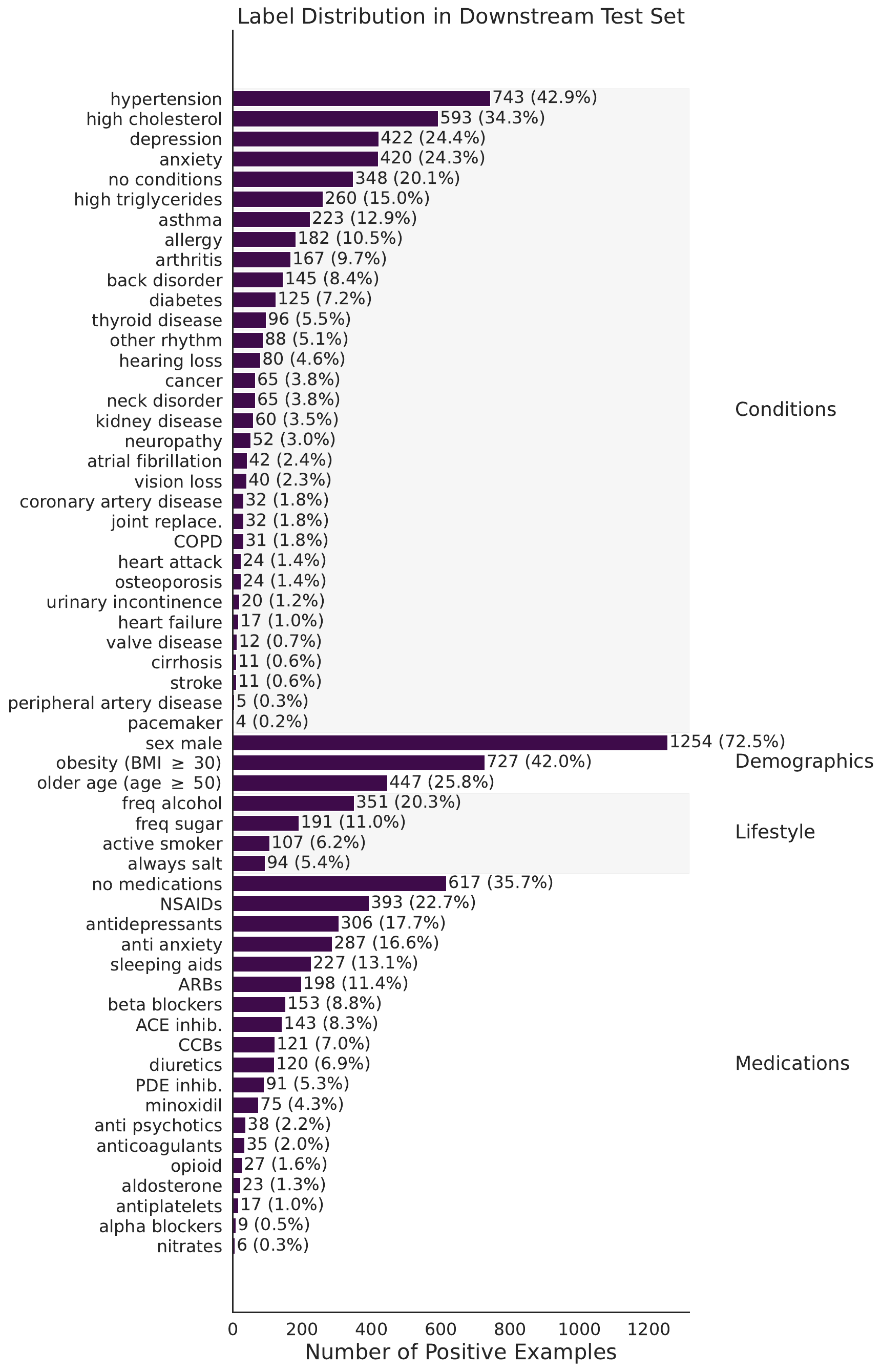}
	\caption{\textbf{Label distribution (downstream test split)}.}
	\label{fig:downstream_label_distribution}
\end{figure*}

\textbf{Downstream dataset.}~The downstream dataset comprises a held-out cohort of $11,736$ participants from the Fitbit Hypertension Study. Data collection was performed under an IRB-approved protocol with informed consent. This dataset includes an average of $11.7$ weeks of recordings per subject, alongside $58$ self-reported features covering demographics, lifestyle behaviors, health conditions, and medical treatments. We performed a subject-level split into training ($70\%$, $N=10,988$), validation ($15\%$, $N=2,356$), and test ($15\%$, $N=2,356$) sets. The demographic and label distributions for the test set are illustrated in \Cref{fig:downstream_demographics} and \Cref{fig:downstream_label_distribution}, respectively. Full text of survey questions for demographics, lifestyle, health conditions, and treatment downstream tasks provided to participants are listed in \Cref{tab:q_demographics} , \Cref{tab:q_lifestyle}, \Cref{tab:q_health_conditions}, \Cref{tab:q_medications}, respectively.

% =========================================================================
% DEMOGRAPHICS TABLE
% =========================================================================
\begin{table}[ht]
\caption{\textbf{Survey Questionnaire: Demographics}}
\vspace{-1.5em}
\label{tab:q_demographics}
\begin{center}
\footnotesize
\begin{tabularx}{0.75\textwidth}{p{5.5cm} X}
\toprule
\textbf{Question} & \textbf{Options / Format} \\
\midrule
What is your date of birth? & Date Picker (e.g., MM/DD/YYYY) \\
\addlinespace
What is your current height? & Text Input (e.g., feet/inches or cm) \\
\addlinespace
What is your current weight? & Text Input (e.g., lbs or kg) \\
\addlinespace
What was your sex assigned at birth? & Male / Female / Other / Prefer not to answer \\
\bottomrule
\end{tabularx}
\end{center}
\end{table}
% =========================================================================
% LIFESTYLE TABLE
% =========================================================================
\begin{table}[h!]
\caption{\textbf{Survey Questionnaire: Lifestyle}}
\vspace{-1.5em}
\label{tab:q_lifestyle}
\begin{center}
\footnotesize
\begin{tabularx}{\textwidth}{p{7cm} X}
\toprule
\textbf{Question} & \textbf{Options} \\
\midrule
How often did you have a drink containing alcohol in the past year? & Never / Monthly or less / Two to four times a month / Two to three times a week / Four or more times a week / I prefer not to answer \\
\addlinespace
Please select the option that best describes your cigarette (or other tobacco products) use: & Never smoker / Former smoker (quit $>5$ years ago) / Former smoker (quit between 1 and 5 years ago) / Former smoker (quit $<1$ year ago or currently using inhaled nicotine doses) / Current smoker / Live with an active smoker in home \\
\addlinespace
Do you add salt to your food? (Do not include salt used in cooking) & Never/rarely / Sometimes / Usually / Always / Prefer not to answer \\
\addlinespace
On average, how many glasses, cans, or cartons containing 250 mL of sugar-sweetened beverages, artificially sweetened beverages, or natural juices do you drink per day? & 0 / More than 0 to 1 / More than 1 / More than 2 / Prefer not to answer \\
\bottomrule
\end{tabularx}
\end{center}
\vspace{-1em}
\end{table}
% =========================================================================
% HEALTH CONDITIONS TABLE
% =========================================================================
\begin{table}[ht]
\caption{\textbf{Survey Questionnaire: Health Conditions Checklist}}
\vspace{-1.5em}
\label{tab:q_health_conditions}
\begin{center}
\footnotesize
\begin{tabularx}{\textwidth}{X}
\toprule
\textbf{Question:} Have you ever been told by a doctor or other health professional that you have or have had any of the following? (\textit{Check all that apply}) \\
\midrule
\begin{minipage}[t]{\linewidth}
\begin{multicols}{2}
\begin{itemize}
    \setlength{\itemsep}{1pt}
    \setlength{\parskip}{0pt}
    \item High blood pressure (hypertension)
    \item Heart attack (myocardial infarction)
    \item Stroke or transient ischemic attack (ministroke)
    \item Atrial Fibrillation (AFib)
    \item Other heart rhythm problem (arrhythmia)
    \item Heart failure (or congestive heart failure)
    \item Coronary artery disease or angina
    \item Peripheral artery disease
    \item Heart valve disease
    \item A pacemaker or an implantable defibrillator (ICD)
    \item Diabetes
    \item High cholesterol
    \item High triglycerides
    \item Arthritis
    \item Hip or knee replacement
    \item Low back disorder / chronic back defect
    \item Neck disorder / chronic neck defect
    \item Osteoporosis
    \item Asthma
    \item Chronic bronchitis, COPD, or emphysema
    \item Rhinitis, hay fever, allergies (excl. asthma)
    \item Kidney problems
    \item Thyroid disease
    \item Cancer
    \item Cirrhosis of the liver
    \item Urinary incontinence
    \item Neuropathy
    \item Depression
    \item Anxiety disorder
    \item Hearing loss
    \item Vision loss
    \item None of the above
\end{itemize}
\end{multicols}
\end{minipage} \\
\bottomrule
\end{tabularx}
\end{center}
\end{table}
% =========================================================================
% MEDICATIONS TABLE
% =========================================================================
\begin{table}[h!]
\caption{\textbf{Survey Questionnaire: Medications Taken in the Past 30 Days}}
\vspace{-1.5em}
\label{tab:q_medications}
\begin{center}
\footnotesize
\begin{tabularx}{\textwidth}{p{4.5cm} X}
\toprule
\multicolumn{2}{p{\textwidth}}{\textbf{Question:} Please check any of the following medications you have taken in the \textbf{\textit{past 30 days.}} (\textit{Note: Examples are provided for each medication class, but are not comprehensive.})} \\
\midrule
\textbf{Medication Class} & \textbf{Examples / Specifics} \\
\midrule
ACE inhibitors & Lisinopril (Zestril, Prinivil), Enalapril (Vasotec), Ramipril (Altace) \\
\addlinespace
Angiotensin Receptor Blockers & Losartan (Cozaar), Valsartan (Diovan), Candesartan (Atacand) \\
\addlinespace
Diuretics & Hydrochlorothiazide (HCTZ) (Microzide), Furosemide (Lasix), Chlorthalidone (Hygroton) \\
\addlinespace
Beta blockers & Metoprolol (Lopressor, Toprol XL), Carvedilol (Coreg), Atenolol (Tenormin) \\
\addlinespace
Calcium channel blockers & Amlodipine (Norvasc), Diltiazem (Cardizem), Verapamil (Calan, Isoptin) \\
\addlinespace
Aldosterone antagonists & Spironolactone (Aldactone), Eplerenone (Inspra) \\
\addlinespace
Minoxidil & Rogaine \\
\addlinespace
Nitrates & Nitroglycerin (Nitrostat, Nitro-Bid), Isosorbide mononitrate (Imdur), Isosorbide dinitrate (Isordil) \\
\addlinespace
Phosphodiesterase inhibitors & Sildenafil (Viagra, Revatio), Tadalafil (Cialis, Adcirca), Vardenafil (Levitra, Staxyn) \\
\addlinespace
Alpha-blockers & Doxazosin (Cardura), Prazosin (Minipress), Terazosin (Hytrin) \\
\addlinespace
Other Classes (\textit{No examples given}) & Anti-anxiety aids, Anti-psychotics, Anticoagulants, Antidepressants, Antiplatelets, Certain types of chemotherapy, Opioid painkillers, NSAIDs, Sleeping aids, or None of the above \\
\bottomrule
\end{tabularx}
\end{center}
\end{table}

\section{Model and Training Details}
\label{sec:implementation_details}

\subsection{\texttt{WavesFM} Implementation Details}
\label{sec:wavesfm_implementation_details}
\textbf{Stage I.}~For the segment encoder $e_\phi$, we employ the EfficientNet-B7 architecture. The input PPG and ACC magnitude segments are treated as $C=2$ distinct channels, yielding an input shape of $C \times l$, where $l = 15\,\text{s} \times 100\,\text{Hz} = 1500$ time steps. This encoder produces segment embeddings $\mathbf{e}_k$ with a dimensionality of $d=256$. The projection head $g_\omega$ is implemented as a two-layer multi-layer perceptron (MLP) that maps these representations to contrastive projections $\mathbf{q}_k$ of dimensionality $d'=128$. During model optimization, we performed a hyperparameter sweep over peak learning rates in the range $[1\text{e-}4, 1\text{e-}2]$ and weight decay values in the range $[1\text{e-}6, 1\text{e-}2]$. The final selected configuration utilizes the Adam optimizer with a peak learning rate of $2\text{e-}4$, a linear warmup of $5,000$ steps, a weight decay of $1\text{e-}6$, and a total training duration of $2,000,000$ steps. The contrastive learning framework is trained with a batch size of $1024$ and a temperature parameter $\gamma = 0.04$.

\textbf{Stage II.}~The temporal encoder consists of a $6$-layer Transformer with a hidden dimension of $d = 256$, an MLP dimension of $768$, and $4$ attention heads. Both the local and global decoders are configured as $4$-layer Transformers with a hidden dimension of $256$ and $4$ attention heads. The local decoder employs target factorized positional embeddings as queries and the encoder context embeddings as keys and values via cross-attention. Conversely, the global decoder processes the concatenation of the mean-pooled context embedding and the target positional embeddings. The factorized positional embeddings are composed of two learnable lookup tables representing the day of week and time of day, denoted as $\mathbf{P}_{\text{dow}} \in \mathbb{R}^{7 \times D}$ and $\mathbf{P}_{\text{tod}} \in \mathbb{R}^{288 \times D}$, respectively. Each input sequence spans one week, divided into $M = 2016$ five-minute bins, from which a context set of size $N_{\text{ctx}} = 252$ is sampled using a patch size $P \sim \text{Uniform}(\{1, 2, 4\})$. Stage II models are trained for $600,000$ iterations with a batch size of $32$ using the Adam optimizer and a cosine decay learning rate schedule, including a linear warmup for the first $1/8$ of the iterations ($75$k steps). While the peak learning rate was searched over the candidate set $\{1\text{e-}4, 4\text{e-}4, 8\text{e-}4\}$, a final learning rate of $8\text{e-}4$ was selected.

\textbf{Compute resources.}~The Stage I segment encoder training was distributed across a cluster equivalent to an NVIDIA \texttt{DGX-2} system ($16\times\text{V100}$ GPUs), with each training run concluding in approximately $96$ hours. The Stage II temporal encoder was optimized using hardware equivalent to an NVIDIA \texttt{V100} GPU, with each training run requiring approximately $24$ hours to complete.

\subsection{Behavior-Centric Baseline Implementation Details}
\label{subsec:behavior_baseline}
To construct a directly comparable behavior-centric baseline, we extract a comprehensive suite of $19$ expert-crafted physiological and kinematic features from the synchronized PPG and ACC signals. Given the raw segments of shape $C \times l = 4 \times 1500$, we compute $10$ PPG-derived cardiac metrics and $9$ ACC-derived kinematic metrics. 

The mathematical formulation and physical descriptions of these $19$ features are cataloged in \Cref{tab:handcrafted_features}. For the cardiac stream, signal cleaning, peak detection, and standard heart rate variability (HRV) metrics are implemented using the \texttt{NeuroKit2}~(\url{https://pypi.org/project/neurokit2/}) library. For the kinematic stream, metrics are designed to capture signal energy, spatial orientation, and movement dynamics. 

\begin{table}[ht]
\centering
\caption{\textbf{Overview of $19$ expert-crafted PPG and ACC features.}}
\label{tab:handcrafted_features}
\footnotesize
\begin{tabular}{llp{9.5cm}}
\toprule
\textbf{Source} & \textbf{Feature Name} & \textbf{Mathematical Description} \\ \midrule
\textbf{PPG} & \texttt{Heart\_Rate\_BPM\_Mean} & Mean heart rate computed across the $15$\,s window in beats per minute. \\
& \texttt{RR\_Mean\_Msec} & Mean interval between successive normal cardiac cycles ($\text{NN}$). \\
& \texttt{RR\_Median\_Msec} & Median interval of the $\text{NN}$ sequence. \\
& \texttt{RR\_20th\_Percentile\_Msec} & 20th percentile value of the $\text{NN}$ interval distribution. \\
& \texttt{RR\_80th\_Percentile\_Msec} & 80th percentile value of the $\text{NN}$ interval distribution. \\
& \texttt{RMSSD\_Msec} & Root mean square of successive differences between adjacent $\text{NN}$ intervals. \\
& \texttt{SDNN\_Msec} & Standard deviation of the $\text{NN}$ intervals. \\
& \texttt{Shannon\_Ent\_RR\_Nats} & Shannon entropy evaluated over the distribution of $\text{NN}$ intervals. \\
& \texttt{Shannon\_Ent\_RR\_Diffs\_Nats} & Shannon entropy of the differences between successive $\text{NN}$ intervals. \\
& \texttt{PNN30\_Percent} & Percentage of successive $\text{NN}$ intervals that differ by more than $30$\,ms. \\ \midrule
\textbf{ACC} & \texttt{Log\_Energy} & Logarithm of the sum of the root-mean-square values across the three axes. \\
& \texttt{Log\_Energy\_Ratio} & Logarithm of the ratio of energy in the first Principal Component ($\text{PC}_1$) to the total 3-axis root-mean-square magnitude energy. \\
& \texttt{Jerk\_Autocorr\_Ratio} & Average lag-$1$ autocorrelation of the kinematic jerk, normalized by the energy of $\text{PC}_1$. \\
& \texttt{Covariance\_Condition} & Matrix condition number of the $3 \times 3$ tri-axial acceleration covariance matrix. \\
& \texttt{Zero\_Crossing\_Avg\_Seconds} & Mean duration between sequential zero-crossings of the $\text{PC}_1$ timeseries. \\
& \texttt{Zero\_Crossing\_Std\_Seconds} & Standard deviation of durations between zero-crossings of the $\text{PC}_1$ timeseries. \\
& \texttt{Robust\_Arm\_Tilt} & Logarithm of the mean root-sum-of-squares of the lateral ($X$) and vertical ($Z$) axes. \\
& \texttt{Kurtosis} & Kurtosis of the 3-axis root-mean-square acceleration magnitude. \\
& \texttt{Sleep\_Coefficient} & Digitized integer assignment of the cumulative peak-to-peak range binned across $16$ log-spaced thresholds. \\
\bottomrule
\end{tabular}
\end{table}

To transform these local features into longitudinal trajectories suitable for temporal modeling, the statistical features extracted from individual windows are aggregated and averaged over $5$-minute resolutions. For a full week-long recording, this mapping converts the underlying high-frequency waveforms into a feature matrix of shape $2016 \times 19$, where $2016$ represents the number of $5$-minute steps in a $7$-day epoch, and $19$ represents the distinct feature dimensionality. 

Prior to model ingestion, we apply independent per-feature standardization (Z-score scaling) across the feature dimensions. This normalized sequence is then passed directly into a Stage II temporal encoder. To ensure a comparable evaluation of representation learning performance, this baseline utilizes an identical architecture configuration ($6$-layer Transformer, $4$ attention heads, $d=256$ hidden state dimensionality) and the exact same hyperparameter schedule detailed in Appendix~\ref{sec:wavesfm_implementation_details}.

\subsection{Supervised Learning Baselines Implementation}
\label{subsec:supervised_learning_implementation}
To contextualize the performance of our self-supervised temporal encoder, we implement a suite of supervised sequence architectures trained directly on the downstream labeled data. For each baseline approach, a standalone sequential model is optimized end-to-end to process the longitudinal sequence of segment embeddings (of shape $M \times d = 2016 \times 256$) and map it directly to the target health label.

\begin{itemize}[leftmargin=*,itemsep=0.4ex,parsep=-0.1ex]
    \item \textbf{Weighted Pooling:}~This architecture maps inputs through an initial \texttt{LayerNorm} block before passing each segment embedding $\mathbf{v}_t \in \mathbb{R}^d$ through a linear gating function $G: \mathbb{R}^d \rightarrow \mathbb{R}$ to compute an unnormalized scalar importance score $s_t$. To handle missingness and irregular sampling gaps, padded elements are masked out by forcing their scores to $-\infty$. The sequence scores are normalized via a softmax operation to yield temporal weights, $\alpha_t = \exp(s_t) / \sum_{j=1}^M \exp(s_j)$, which are used to compress the sequence into a static representation, $\bar{\mathbf{v}} = \sum_{t=1}^M \alpha_t \mathbf{v}_t$. This pooled context vector is passed through a dropout layer ($\rho=0.2$) and evaluated by a final linear dense projection head.
    \item \textbf{Cross-Attention:}~This model computes directed interactions between a learnable classification token ($\mathbf{q}_{\text{CLS}} \in \mathbb{R}^d$) and the sequence context. The $\text{CLS}$ token serves as the query ($\mathbf{Q}$), while the normalized longitudinal embeddings function as keys ($\mathbf{K}$) and values ($\mathbf{V}$). Absolute positional relationships are preserved by factoring a Rotary Positional Embedding (RoPE) \cite{su2024roformer} directly into the query and key heads based on their actual index positions (with the query evaluated at position $0$). An asymmetric boolean mask restricts the query token from attending to invalid or missing context intervals. The updated classification embedding is routed through a standard two-layer Multi-Layer Perceptron (MLP) block utilizing a $\text{GELU}$ activation layer and dropout ($\rho=0.2$) before emitting the final projection logits.
    \item \textbf{LSTM:}~This model captures long-term autoregressive dependencies using a recurrent Long Short-Term Memory network \cite{hochreiter1997long} configured with a hidden dimensionality of $64$. To prevent invalid intervals or missing segments from corrupting the forward memory cell parameters, the model evaluates an element-wise conditional check; if a step fails the missingness check, the cell update step propagates the un-modified hidden state ($\mathbf{h}_{t-1}$) and cell state ($\mathbf{c}_{t-1}$) directly into the subsequent window step. The terminal hidden state ($\mathbf{h}_M$) is filtered through a dropout layer ($\rho=0.2$) and processed by a dense linear projection head to yield the final classification logit.
\end{itemize}

\subsection{Evaluating Open-Source PPG Foundation Models}
To assess the representation capability of our foundation model, we benchmark downstream performance against a collection of open-source and pretrained PPG foundation models. These models represent the latest state-of-the-art in self-supervised and cross-modal pretraining for PPG, providing critical reference points to evaluate generalizability. Specifically, we include \texttt{PaPaGei} \cite{pillai2025papagei}, \texttt{Pulse-PPG} \cite{saha2025pulse}, and \texttt{AnyPPG} \cite{nie2025anyppg} in our comparative analysis.

The details of these three state-of-the-art baselines are detailed as follows:
\vspace{-1em}
\begin{itemize}[leftmargin=*,itemsep=0.4ex,parsep=-0.1ex]
    \item \textbf{PaPaGei:} this model utilizes a ResNet optimized through a morphology-aware framework that pairs signals with similar blood volume dynamics across different participants and simultaneously trains multiple expert heads to predict key physiological pulse shape metrics. The encoder processes $10$s segments sampled at $125$Hz, producing a static representation vector of dimensionality $d=512$ per segment.
    \item \textbf{Pulse-PPG:} Pulse-PPG utilizes a 1D ResNet-26 backbone optimized via a motif-based relative contrastive learning framework, explicitly mapping the non-linear distances between waveform segments. The model captures long-term contextual patterns over $4$-minute window inputs resampled at $50$Hz, ultimately transforming each signal segment into a vector of dimensionality $d=512$.
    \item \textbf{AnyPPG:} AnyPPG shifts the unimodal paradigm by executing a cross-modal, dual-branch pretraining objective where a 1D ResNet PPG encoder is guided symmetrically via text-image alignment metrics alongside a frozen \texttt{ECGFounder} cross-modal backbone. Trained over a heterogeneous multi-source corpus, the PPG branch evaluates filtered $10$s signal slices at $125$\,Hz to map localized cardiovascular dynamics directly into a final embedding space of dimensionality $d=512$.
\end{itemize}

To ensure a fair comparison, raw PPG waveforms are independently preprocessed, resampled, and structurally framed to match the input settings specified by each respective open-source baseline. For each isolated segment extracted from a given dataset, we retrieve the corresponding frozen embedding vector ($d=512$ for all three models). These segment-level representations are subsequently averaged across the temporal domain at the subject level to derive a single subject-specific representation vector, which is then utilized to train and evaluate downstream linear probes.
\section{Detailed Results}
\label{sec:detailed-results}

\begin{figure*}[h!]
\vspace{-0.5em}
	\centering
	\includegraphics[width=0.9\linewidth, keepaspectratio]{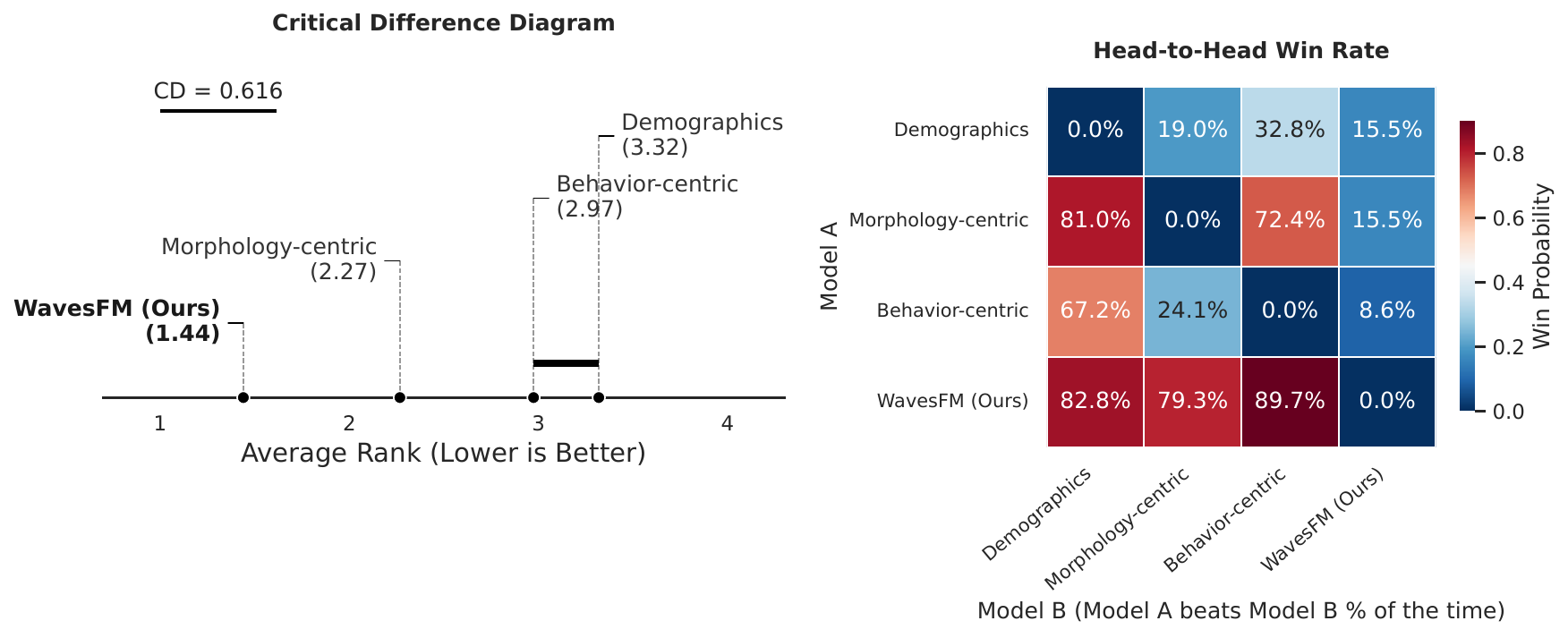}
    \vspace{-0.5em}
	\caption{\textbf{Statistical evaluation across 58 benchmark tasks.}~(Left) Critical Difference diagram based on post-hoc Nemenyi test ($\alpha$ = 0.05, CD = 0.616). \texttt{WavesFM} achieves a significantly better average rank (1.44) than all baselines; thick horizontal bars indicate no statistically significant difference ($p \geq 0.05$). (Right) Pairwise win rate matrix showing the percentage of tasks where Model A yields a superior AUROC over Model B. \texttt{WavesFM} demonstrates robust multi-task performance.}
	\label{fig:stat_ranking}
    \vspace{-1em}

\end{figure*}

\subsection{Comparison with Morphology- and Behavior-Centric Baselines}
  In \Cref{tab:detailed_results}, we present the comprehensive performance metrics across all classification tasks. All reported metrics include 95\% confidence intervals derived from 1000 test-set bootstrap iterations. Additionally, we provide the mean AUC and pAUC aggregated by category, alongside the overall mean across all tasks. \texttt{WavesFM} achieves the highest performance on the majority of tasks, with the best results highlighted in bold.

We systematically evaluate the generalization capabilities of our framework across the complete 58-task benchmark suite. The Critical Difference diagram (Figure \ref{fig:stat_ranking}) shows that WavesFM achieved the top average rank (1.44) and is statistically distinct from all baselines. This robust generalization is further emphasized by the head-to-head win matrix.

\begin{table}[p]
    \centering
    \captionsetup{font=footnotesize}
    \caption{\textbf{Detailed Results for Downstream Tasks.} CCB: calcium channel blocker; ARB: angiotension receptor blocker; PDE: phosphodiesterase; ACE: angiotension-converting enzyme; NSAID: nonsteroidal anti-inflammatory drugs; PAD: peripheral artery disease; CAD: coronary artery disease; AFib: atrial fibrillation; COPD: chronic obstructive pulmonary disease}
    \label{tab:detailed_results}
    \renewcommand{\arraystretch}{1.2}
    \resizebox{\textwidth}{!}{
    \begin{tabular}{l cccccccc}
        \toprule
        \multirow{2}{*}{\textbf{Task}} & \multicolumn{2}{c}{\textbf{Demographics}} & \multicolumn{2}{c}{\textbf{Behavior-Centric}} & \multicolumn{2}{c}{\textbf{Morphology-Centric}} & \multicolumn{2}{c}{\textbf{\texttt{WavesFM}}} \\
        \cmidrule(lr){2-3} \cmidrule(lr){4-5} \cmidrule(lr){6-7} \cmidrule(lr){8-9}
        & AUROC$^\uparrow$ & pAUC$^\uparrow$ & AUROC$^\uparrow$ & pAUC$^\uparrow$ & AUROC$^\uparrow$ & pAUC$^\uparrow$ & AUROC$^\uparrow$ & pAUC$^\uparrow$ \\
        \midrule
        \multicolumn{9}{l}{\textit{\textbf{Demographics}}} \\
        sex male & 1.000 \scriptsize{[1.000, 1.000]} & 1.000 \scriptsize{[1.000, 1.000]} & 0.975 \scriptsize{[0.968, 0.981]} & 0.913 \scriptsize{[0.889, 0.938]} & 0.993 \scriptsize{[0.988, 0.996]} & 0.971 \scriptsize{[0.947, 0.986]} & \textbf{0.996 \scriptsize{[0.992, 0.998]}} & \textbf{0.983 \scriptsize{[0.962, 0.994]}} \\
        older age (age $\ge$ 50) & 1.000 \scriptsize{[1.000, 1.000]} & 1.000 \scriptsize{[1.000, 1.000]} & 0.881 \scriptsize{[0.863, 0.897]} & 0.724 \scriptsize{[0.697, 0.750]} & 0.945 \scriptsize{[0.934, 0.955]} & 0.824 \scriptsize{[0.799, 0.853]} & \textbf{0.959 \scriptsize{[0.950, 0.968]}} & \textbf{0.852 \scriptsize{[0.829, 0.877]}} \\
        obesity (BMI $\ge$ 30) & 1.000 \scriptsize{[1.000, 1.000]} & 1.000 \scriptsize{[1.000, 1.000]} & 0.817 \scriptsize{[0.796, 0.837]} & 0.663 \scriptsize{[0.641, 0.688]} & 0.921 \scriptsize{[0.908, 0.933]} & 0.798 \scriptsize{[0.773, 0.824]} & \textbf{0.928 \scriptsize{[0.917, 0.940]}} & \textbf{0.814 \scriptsize{[0.790, 0.837]}} \\
        \rowcolor[gray]{.95} \textbf{Mean (Demographics)} & 1.000 \scriptsize{$\pm$ 0.000} & 1.000 \scriptsize{$\pm$ 0.000} & 0.891 \scriptsize{$\pm$ 0.079} & 0.767 \scriptsize{$\pm$ 0.131} & 0.953 \scriptsize{$\pm$ 0.037} & 0.864 \scriptsize{$\pm$ 0.093} & \textbf{0.961 \scriptsize{$\pm$ 0.034}} & \textbf{0.883 \scriptsize{$\pm$ 0.089}} \\
        \midrule
        \addlinespace
        \multicolumn{9}{l}{\textit{\textbf{Lifestyle}}} \\
        freq sugar & 0.583 \scriptsize{[0.554, 0.611]} & 0.512 \scriptsize{[0.501, 0.525]} & 0.694 \scriptsize{[0.654, 0.733]} & 0.556 \scriptsize{[0.533, 0.583]} & 0.688 \scriptsize{[0.650, 0.726]} & 0.559 \scriptsize{[0.535, 0.584]} & \textbf{0.697 \scriptsize{[0.658, 0.736]}} & \textbf{0.561 \scriptsize{[0.538, 0.587]}} \\
        always salt & 0.613 \scriptsize{[0.566, 0.660]} & 0.524 \scriptsize{[0.502, 0.549]} & 0.598 \scriptsize{[0.540, 0.655]} & 0.514 \scriptsize{[0.491, 0.541]} & 0.598 \scriptsize{[0.540, 0.652]} & 0.520 \scriptsize{[0.496, 0.546]} & \textbf{0.605 \scriptsize{[0.545, 0.660]}} & \textbf{0.524 \scriptsize{[0.499, 0.550]}} \\
        active smoker & 0.612 \scriptsize{[0.568, 0.654]} & 0.525 \scriptsize{[0.503, 0.548]} & 0.722 \scriptsize{[0.664, 0.776]} & 0.580 \scriptsize{[0.546, 0.616]} & 0.736 \scriptsize{[0.686, 0.787]} & 0.603 \scriptsize{[0.569, 0.639]} & \textbf{0.784 \scriptsize{[0.734, 0.830]}} & \textbf{0.635 \scriptsize{[0.596, 0.677]}} \\
        freq alcohol & 0.583 \scriptsize{[0.541, 0.623]} & 0.539 \scriptsize{[0.521, 0.560]} & 0.743 \scriptsize{[0.714, 0.774]} & 0.583 \scriptsize{[0.561, 0.608]} & 0.661 \scriptsize{[0.630, 0.692]} & 0.540 \scriptsize{[0.523, 0.559]} & \textbf{0.833 \scriptsize{[0.808, 0.857]}} & \textbf{0.672 \scriptsize{[0.642, 0.703]}} \\
        \rowcolor[gray]{.95} \textbf{Mean (Lifestyle)} & 0.598 \scriptsize{$\pm$ 0.017} & 0.525 \scriptsize{$\pm$ 0.011} & 0.689 \scriptsize{$\pm$ 0.064} & 0.558 \scriptsize{$\pm$ 0.032} & 0.671 \scriptsize{$\pm$ 0.058} & 0.555 \scriptsize{$\pm$ 0.036} & \textbf{0.730 \scriptsize{$\pm$ 0.100}} & \textbf{0.598 \scriptsize{$\pm$ 0.068}} \\
        \midrule
        \addlinespace
        \multicolumn{9}{l}{\textit{\textbf{Treatment}}} \\
        nitrates & 0.756 \scriptsize{[0.620, 0.860]} & 0.499 \scriptsize{[0.474, 0.562]} & 0.805 \scriptsize{[0.692, 0.913]} & 0.559 \scriptsize{[0.474, 0.735]} & 0.811 \scriptsize{[0.683, 0.928]} & 0.577 \scriptsize{[0.474, 0.717]} & \textbf{0.864 \scriptsize{[0.770, 0.940]}} & \textbf{0.596 \scriptsize{[0.474, 0.746]}} \\
        antiplatelets & 0.735 \scriptsize{[0.607, 0.871]} & 0.658 \scriptsize{[0.551, 0.748]} & 0.816 \scriptsize{[0.701, 0.910]} & 0.678 \scriptsize{[0.570, 0.783]} & 0.833 \scriptsize{[0.717, 0.940]} & \textbf{0.731 \scriptsize{[0.621, 0.849]}} & \textbf{0.846 \scriptsize{[0.743, 0.937]}} & 0.719 \scriptsize{[0.606, 0.834]} \\
        diuretics & 0.753 \scriptsize{[0.710, 0.790]} & 0.579 \scriptsize{[0.549, 0.603]} & 0.727 \scriptsize{[0.678, 0.771]} & 0.570 \scriptsize{[0.538, 0.602]} & 0.778 \scriptsize{[0.738, 0.817]} & 0.600 \scriptsize{[0.567, 0.637]} & \textbf{0.788 \scriptsize{[0.749, 0.828]}} & \textbf{0.607 \scriptsize{[0.573, 0.643]}} \\
        CCBs & 0.703 \scriptsize{[0.660, 0.742]} & 0.561 \scriptsize{[0.534, 0.588]} & 0.691 \scriptsize{[0.643, 0.740]} & 0.564 \scriptsize{[0.533, 0.596]} & 0.779 \scriptsize{[0.731, 0.824]} & 0.640 \scriptsize{[0.601, 0.684]} & \textbf{0.801 \scriptsize{[0.756, 0.843]}} & \textbf{0.664 \scriptsize{[0.625, 0.708]}} \\
        anticoagulants & 0.710 \scriptsize{[0.633, 0.775]} & 0.558 \scriptsize{[0.513, 0.605]} & 0.770 \scriptsize{[0.673, 0.854]} & 0.614 \scriptsize{[0.551, 0.680]} & \textbf{0.778 \scriptsize{[0.693, 0.852]}} & \textbf{0.620 \scriptsize{[0.559, 0.683]}} & 0.772 \scriptsize{[0.683, 0.850]} & 0.617 \scriptsize{[0.557, 0.680]} \\
        beta blockers & 0.667 \scriptsize{[0.627, 0.706]} & 0.551 \scriptsize{[0.527, 0.572]} & 0.752 \scriptsize{[0.708, 0.791]} & 0.616 \scriptsize{[0.582, 0.650]} & 0.781 \scriptsize{[0.737, 0.818]} & 0.637 \scriptsize{[0.601, 0.671]} & \textbf{0.807 \scriptsize{[0.763, 0.846]}} & \textbf{0.664 \scriptsize{[0.627, 0.700]}} \\
        ARBs & 0.728 \scriptsize{[0.697, 0.758]} & 0.560 \scriptsize{[0.538, 0.583]} & 0.676 \scriptsize{[0.637, 0.715]} & 0.551 \scriptsize{[0.527, 0.577]} & 0.737 \scriptsize{[0.702, 0.771]} & 0.580 \scriptsize{[0.553, 0.608]} & \textbf{0.753 \scriptsize{[0.719, 0.785]}} & \textbf{0.589 \scriptsize{[0.562, 0.618]}} \\
        PDE inhib. & \textbf{0.744 \scriptsize{[0.708, 0.778]}} & 0.537 \scriptsize{[0.513, 0.565]} & 0.716 \scriptsize{[0.659, 0.770]} & \textbf{0.561 \scriptsize{[0.527, 0.596]}} & 0.735 \scriptsize{[0.691, 0.777]} & 0.546 \scriptsize{[0.516, 0.578]} & 0.731 \scriptsize{[0.686, 0.775]} & 0.548 \scriptsize{[0.517, 0.585]} \\
        ACE inhib. & 0.723 \scriptsize{[0.687, 0.757]} & 0.558 \scriptsize{[0.535, 0.582]} & 0.690 \scriptsize{[0.644, 0.732]} & 0.540 \scriptsize{[0.515, 0.566]} & \textbf{0.727 \scriptsize{[0.687, 0.766]}} & 0.567 \scriptsize{[0.539, 0.600]} & 0.723 \scriptsize{[0.686, 0.762]} & \textbf{0.571 \scriptsize{[0.543, 0.605]}} \\
        aldosterone & 0.666 \scriptsize{[0.568, 0.757]} & 0.508 \scriptsize{[0.474, 0.549]} & 0.738 \scriptsize{[0.623, 0.834]} & 0.559 \scriptsize{[0.503, 0.621]} & \textbf{0.762 \scriptsize{[0.651, 0.851]}} & \textbf{0.587 \scriptsize{[0.521, 0.658]}} & 0.729 \scriptsize{[0.622, 0.813]} & 0.543 \scriptsize{[0.490, 0.604]} \\
        alpha blockers & 0.642 \scriptsize{[0.504, 0.754]} & 0.474 \scriptsize{[0.474, 0.475]} & 0.702 \scriptsize{[0.521, 0.826]} & 0.474 \scriptsize{[0.474, 0.477]} & 0.666 \scriptsize{[0.521, 0.796]} & 0.477 \scriptsize{[0.474, 0.494]} & \textbf{0.710 \scriptsize{[0.536, 0.860]}} & \textbf{0.483 \scriptsize{[0.474, 0.513]}} \\
        minoxidil & 0.652 \scriptsize{[0.597, 0.703]} & 0.529 \scriptsize{[0.505, 0.556]} & 0.629 \scriptsize{[0.568, 0.691]} & 0.536 \scriptsize{[0.505, 0.568]} & 0.663 \scriptsize{[0.598, 0.725]} & 0.541 \scriptsize{[0.512, 0.576]} & \textbf{0.687 \scriptsize{[0.627, 0.747]}} & \textbf{0.565 \scriptsize{[0.529, 0.607]}} \\
        opioid & 0.631 \scriptsize{[0.536, 0.711]} & 0.529 \scriptsize{[0.498, 0.567]} & 0.786 \scriptsize{[0.683, 0.881]} & 0.660 \scriptsize{[0.586, 0.733]} & 0.785 \scriptsize{[0.683, 0.888]} & 0.719 \scriptsize{[0.638, 0.805]} & \textbf{0.831 \scriptsize{[0.727, 0.916]}} & \textbf{0.749 \scriptsize{[0.656, 0.839]}} \\
        NSAIDs & 0.580 \scriptsize{[0.553, 0.607]} & 0.517 \scriptsize{[0.505, 0.529]} & 0.581 \scriptsize{[0.549, 0.612]} & 0.519 \scriptsize{[0.506, 0.534]} & \textbf{0.619 \scriptsize{[0.589, 0.650]}} & \textbf{0.520 \scriptsize{[0.506, 0.534]}} & 0.615 \scriptsize{[0.585, 0.648]} & \textbf{0.520 \scriptsize{[0.506, 0.536]}} \\
        antidepressants & 0.611 \scriptsize{[0.581, 0.642]} & 0.531 \scriptsize{[0.516, 0.547]} & 0.741 \scriptsize{[0.710, 0.771]} & 0.595 \scriptsize{[0.572, 0.621]} & 0.771 \scriptsize{[0.742, 0.798]} & 0.617 \scriptsize{[0.591, 0.642]} & \textbf{0.819 \scriptsize{[0.793, 0.846]}} & \textbf{0.664 \scriptsize{[0.634, 0.695]}} \\
        anti psychotics & 0.577 \scriptsize{[0.507, 0.638]} & 0.507 \scriptsize{[0.483, 0.538]} & 0.790 \scriptsize{[0.703, 0.869]} & 0.658 \scriptsize{[0.593, 0.724]} & 0.752 \scriptsize{[0.670, 0.827]} & 0.602 \scriptsize{[0.548, 0.660]} & \textbf{0.828 \scriptsize{[0.751, 0.898]}} & \textbf{0.681 \scriptsize{[0.618, 0.750]}} \\
        anti anxiety & 0.615 \scriptsize{[0.584, 0.645]} & 0.530 \scriptsize{[0.517, 0.548]} & 0.721 \scriptsize{[0.687, 0.752]} & 0.576 \scriptsize{[0.551, 0.601]} & 0.724 \scriptsize{[0.692, 0.754]} & 0.571 \scriptsize{[0.547, 0.595]} & \textbf{0.757 \scriptsize{[0.724, 0.787]}} & \textbf{0.604 \scriptsize{[0.580, 0.633]}} \\
        sleeping aids & 0.563 \scriptsize{[0.531, 0.599]} & 0.513 \scriptsize{[0.500, 0.529]} & 0.634 \scriptsize{[0.592, 0.673]} & 0.535 \scriptsize{[0.517, 0.555]} & 0.605 \scriptsize{[0.567, 0.645]} & 0.531 \scriptsize{[0.512, 0.553]} & \textbf{0.642 \scriptsize{[0.604, 0.683]}} & \textbf{0.549 \scriptsize{[0.527, 0.572]}} \\
        none & 0.624 \scriptsize{[0.601, 0.648]} & 0.531 \scriptsize{[0.519, 0.545]} & 0.669 \scriptsize{[0.644, 0.694]} & 0.542 \scriptsize{[0.526, 0.558]} & 0.710 \scriptsize{[0.688, 0.735]} & \textbf{0.553 \scriptsize{[0.537, 0.575]}} & \textbf{0.720 \scriptsize{[0.697, 0.743]}} & 0.549 \scriptsize{[0.532, 0.571]} \\
        \rowcolor[gray]{.95} \textbf{Mean (Treatment)} & 0.667 \scriptsize{$\pm$ 0.063} & 0.538 \scriptsize{$\pm$ 0.039} & 0.718 \scriptsize{$\pm$ 0.063} & 0.574 \scriptsize{$\pm$ 0.052} & 0.738 \scriptsize{$\pm$ 0.062} & 0.590 \scriptsize{$\pm$ 0.063} & \textbf{0.759 \scriptsize{$\pm$ 0.068}} & \textbf{0.604 \scriptsize{$\pm$ 0.070}} \\
        \midrule
        \addlinespace
        \multicolumn{9}{l}{\textit{\textbf{Health Conditions}}} \\
        cancer & \textbf{0.741 \scriptsize{[0.682, 0.792]}} & \textbf{0.594 \scriptsize{[0.550, 0.633]}} & 0.673 \scriptsize{[0.596, 0.741]} & 0.574 \scriptsize{[0.530, 0.622]} & 0.708 \scriptsize{[0.638, 0.773]} & 0.574 \scriptsize{[0.531, 0.619]} & 0.718 \scriptsize{[0.651, 0.781]} & 0.583 \scriptsize{[0.540, 0.626]} \\
        PAD & 0.902 \scriptsize{[0.785, 0.974]} & 0.709 \scriptsize{[0.534, 0.864]} & 0.973 \scriptsize{[0.923, 0.998]} & 0.870 \scriptsize{[0.645, 0.991]} & \textbf{0.977 \scriptsize{[0.929, 1.000]}} & \textbf{0.887 \scriptsize{[0.648, 1.000]}} & 0.968 \scriptsize{[0.899, 1.000]} & 0.884 \scriptsize{[0.647, 1.000]} \\
        pacemaker & 0.710 \scriptsize{[0.544, 0.895]} & 0.547 \scriptsize{[0.474, 0.726]} & 0.935 \scriptsize{[0.885, 0.999]} & 0.676 \scriptsize{[0.474, 0.995]} & 0.925 \scriptsize{[0.738, 1.000]} & \textbf{0.843 \scriptsize{[0.474, 1.000]}} & \textbf{0.969 \scriptsize{[0.913, 1.000]}} & 0.837 \scriptsize{[0.547, 1.000]} \\
        CAD & 0.821 \scriptsize{[0.768, 0.868]} & 0.613 \scriptsize{[0.560, 0.674]} & 0.830 \scriptsize{[0.750, 0.901]} & 0.671 \scriptsize{[0.601, 0.746]} & 0.866 \scriptsize{[0.796, 0.924]} & \textbf{0.688 \scriptsize{[0.610, 0.771]}} & \textbf{0.871 \scriptsize{[0.811, 0.923]}} & 0.686 \scriptsize{[0.607, 0.770]} \\
        heart failure & 0.771 \scriptsize{[0.681, 0.862]} & 0.574 \scriptsize{[0.509, 0.651]} & 0.797 \scriptsize{[0.637, 0.909]} & 0.625 \scriptsize{[0.531, 0.731]} & 0.832 \scriptsize{[0.711, 0.941]} & \textbf{0.712 \scriptsize{[0.598, 0.827]}} & \textbf{0.882 \scriptsize{[0.808, 0.949]}} & \textbf{0.712 \scriptsize{[0.611, 0.835]}} \\
        AFib & 0.813 \scriptsize{[0.761, 0.864]} & 0.630 \scriptsize{[0.579, 0.683]} & 0.791 \scriptsize{[0.720, 0.861]} & 0.624 \scriptsize{[0.564, 0.693]} & \textbf{0.823 \scriptsize{[0.754, 0.887]}} & 0.658 \scriptsize{[0.597, 0.721]} & \textbf{0.823 \scriptsize{[0.753, 0.887]}} & \textbf{0.664 \scriptsize{[0.600, 0.730]}} \\
        heart attack & 0.744 \scriptsize{[0.654, 0.828]} & 0.595 \scriptsize{[0.534, 0.661]} & 0.832 \scriptsize{[0.753, 0.900]} & 0.639 \scriptsize{[0.558, 0.727]} & 0.852 \scriptsize{[0.765, 0.925]} & \textbf{0.677 \scriptsize{[0.595, 0.766]}} & \textbf{0.860 \scriptsize{[0.781, 0.930]}} & 0.669 \scriptsize{[0.584, 0.767]} \\
        hypertension & 0.689 \scriptsize{[0.668, 0.711]} & 0.559 \scriptsize{[0.544, 0.573]} & 0.694 \scriptsize{[0.668, 0.722]} & 0.560 \scriptsize{[0.542, 0.581]} & 0.734 \scriptsize{[0.711, 0.758]} & 0.594 \scriptsize{[0.576, 0.613]} & \textbf{0.739 \scriptsize{[0.717, 0.763]}} & \textbf{0.597 \scriptsize{[0.578, 0.618]}} \\
        stroke & 0.599 \scriptsize{[0.445, 0.752]} & 0.606 \scriptsize{[0.515, 0.708]} & 0.623 \scriptsize{[0.460, 0.780]} & 0.512 \scriptsize{[0.474, 0.605]} & 0.667 \scriptsize{[0.491, 0.829]} & 0.542 \scriptsize{[0.474, 0.659]} & \textbf{0.736 \scriptsize{[0.582, 0.878]}} & \textbf{0.602 \scriptsize{[0.501, 0.725]}} \\
        other rhythm & 0.591 \scriptsize{[0.542, 0.643]} & 0.504 \scriptsize{[0.487, 0.523]} & \textbf{0.674 \scriptsize{[0.616, 0.728]}} & 0.535 \scriptsize{[0.506, 0.570]} & 0.636 \scriptsize{[0.574, 0.695]} & 0.535 \scriptsize{[0.506, 0.568]} & 0.660 \scriptsize{[0.601, 0.715]} & \textbf{0.539 \scriptsize{[0.510, 0.574]}} \\
        valve disease & 0.551 \scriptsize{[0.426, 0.713]} & 0.513 \scriptsize{[0.474, 0.575]} & 0.615 \scriptsize{[0.406, 0.819]} & \textbf{0.573 \scriptsize{[0.489, 0.677]}} & \textbf{0.671 \scriptsize{[0.510, 0.816]}} & 0.541 \scriptsize{[0.474, 0.661]} & 0.643 \scriptsize{[0.485, 0.798]} & 0.529 \scriptsize{[0.474, 0.635]} \\
        cirrhosis & 0.607 \scriptsize{[0.444, 0.765]} & 0.559 \scriptsize{[0.492, 0.640]} & 0.637 \scriptsize{[0.472, 0.801]} & 0.557 \scriptsize{[0.474, 0.672]} & \textbf{0.725 \scriptsize{[0.590, 0.856]}} & 0.592 \scriptsize{[0.474, 0.720]} & 0.689 \scriptsize{[0.557, 0.832]} & \textbf{0.607 \scriptsize{[0.474, 0.754]}} \\
        anxiety & 0.680 \scriptsize{[0.656, 0.705]} & 0.571 \scriptsize{[0.553, 0.592]} & 0.756 \scriptsize{[0.731, 0.781]} & 0.599 \scriptsize{[0.574, 0.625]} & 0.763 \scriptsize{[0.737, 0.788]} & 0.601 \scriptsize{[0.579, 0.624]} & \textbf{0.790 \scriptsize{[0.765, 0.814]}} & \textbf{0.635 \scriptsize{[0.612, 0.662]}} \\
        depression & 0.657 \scriptsize{[0.631, 0.681]} & 0.552 \scriptsize{[0.537, 0.570]} & 0.726 \scriptsize{[0.697, 0.755]} & 0.591 \scriptsize{[0.570, 0.614]} & 0.746 \scriptsize{[0.719, 0.771]} & 0.598 \scriptsize{[0.577, 0.622]} & \textbf{0.771 \scriptsize{[0.745, 0.797]}} & \textbf{0.624 \scriptsize{[0.601, 0.651]}} \\
        diabetes & 0.738 \scriptsize{[0.699, 0.773]} & 0.575 \scriptsize{[0.548, 0.602]} & 0.796 \scriptsize{[0.755, 0.836]} & 0.631 \scriptsize{[0.595, 0.670]} & 0.816 \scriptsize{[0.776, 0.852]} & 0.656 \scriptsize{[0.615, 0.695]} & \textbf{0.836 \scriptsize{[0.796, 0.876]}} & \textbf{0.691 \scriptsize{[0.649, 0.731]}} \\
        thyroid disease & 0.720 \scriptsize{[0.675, 0.767]} & \textbf{0.593 \scriptsize{[0.563, 0.626]}} & 0.710 \scriptsize{[0.654, 0.762]} & 0.551 \scriptsize{[0.522, 0.586]} & 0.719 \scriptsize{[0.665, 0.769]} & 0.556 \scriptsize{[0.524, 0.593]} & \textbf{0.722 \scriptsize{[0.667, 0.773]}} & 0.560 \scriptsize{[0.528, 0.595]} \\
        high cholesterol & 0.631 \scriptsize{[0.607, 0.654]} & \textbf{0.537 \scriptsize{[0.524, 0.551]}} & 0.614 \scriptsize{[0.586, 0.640]} & 0.530 \scriptsize{[0.516, 0.546]} & 0.640 \scriptsize{[0.613, 0.664]} & 0.525 \scriptsize{[0.511, 0.540]} & \textbf{0.647 \scriptsize{[0.620, 0.672]}} & 0.533 \scriptsize{[0.519, 0.548]} \\
        high triglycerides & 0.598 \scriptsize{[0.572, 0.630]} & 0.504 \scriptsize{[0.494, 0.516]} & 0.620 \scriptsize{[0.585, 0.656]} & \textbf{0.526 \scriptsize{[0.509, 0.545]}} & 0.607 \scriptsize{[0.568, 0.640]} & 0.521 \scriptsize{[0.504, 0.540]} & \textbf{0.625 \scriptsize{[0.589, 0.657]}} & 0.525 \scriptsize{[0.508, 0.542]} \\
        joint replace. & 0.831 \scriptsize{[0.765, 0.884]} & 0.641 \scriptsize{[0.582, 0.701]} & 0.810 \scriptsize{[0.747, 0.874]} & 0.636 \scriptsize{[0.567, 0.713]} & \textbf{0.866 \scriptsize{[0.813, 0.913]}} & 0.655 \scriptsize{[0.584, 0.724]} & \textbf{0.866 \scriptsize{[0.810, 0.915]}} & \textbf{0.664 \scriptsize{[0.594, 0.738]}} \\
        arthritis & 0.758 \scriptsize{[0.726, 0.788]} & 0.585 \scriptsize{[0.561, 0.613]} & 0.763 \scriptsize{[0.722, 0.798]} & 0.590 \scriptsize{[0.561, 0.620]} & 0.795 \scriptsize{[0.762, 0.826]} & 0.607 \scriptsize{[0.576, 0.639]} & \textbf{0.802 \scriptsize{[0.770, 0.831]}} & \textbf{0.608 \scriptsize{[0.576, 0.642]}} \\
        osteoporosis & \textbf{0.804 \scriptsize{[0.706, 0.882]}} & \textbf{0.701 \scriptsize{[0.627, 0.773]}} & 0.767 \scriptsize{[0.656, 0.865]} & 0.615 \scriptsize{[0.532, 0.698]} & 0.763 \scriptsize{[0.637, 0.869]} & 0.660 \scriptsize{[0.568, 0.749]} & 0.767 \scriptsize{[0.636, 0.877]} & 0.651 \scriptsize{[0.561, 0.740]} \\
        neck disorder & 0.672 \scriptsize{[0.614, 0.728]} & 0.535 \scriptsize{[0.509, 0.567]} & 0.649 \scriptsize{[0.578, 0.716]} & 0.544 \scriptsize{[0.507, 0.586]} & 0.699 \scriptsize{[0.628, 0.762]} & 0.572 \scriptsize{[0.529, 0.615]} & \textbf{0.705 \scriptsize{[0.638, 0.766]}} & \textbf{0.576 \scriptsize{[0.532, 0.622]}} \\
        back disorder & 0.634 \scriptsize{[0.596, 0.673]} & 0.522 \scriptsize{[0.504, 0.540]} & 0.670 \scriptsize{[0.622, 0.712]} & 0.548 \scriptsize{[0.523, 0.574]} & 0.687 \scriptsize{[0.641, 0.733]} & 0.558 \scriptsize{[0.531, 0.587]} & \textbf{0.703 \scriptsize{[0.658, 0.750]}} & \textbf{0.576 \scriptsize{[0.547, 0.609]}} \\
        neuropathy & 0.697 \scriptsize{[0.637, 0.749]} & 0.539 \scriptsize{[0.505, 0.566]} & 0.726 \scriptsize{[0.646, 0.800]} & 0.584 \scriptsize{[0.537, 0.635]} & \textbf{0.770 \scriptsize{[0.695, 0.837]}} & 0.646 \scriptsize{[0.588, 0.706]} & \textbf{0.770 \scriptsize{[0.696, 0.840]}} & \textbf{0.653 \scriptsize{[0.590, 0.715]}} \\
        kidney disease & 0.620 \scriptsize{[0.563, 0.691]} & 0.548 \scriptsize{[0.511, 0.578]} & 0.644 \scriptsize{[0.558, 0.723]} & 0.574 \scriptsize{[0.533, 0.620]} & 0.644 \scriptsize{[0.561, 0.725]} & 0.580 \scriptsize{[0.538, 0.634]} & \textbf{0.688 \scriptsize{[0.611, 0.759]}} & \textbf{0.591 \scriptsize{[0.545, 0.644]}} \\
        COPD & 0.596 \scriptsize{[0.490, 0.696]} & 0.518 \scriptsize{[0.483, 0.560]} & \textbf{0.749 \scriptsize{[0.676, 0.813]}} & 0.539 \scriptsize{[0.493, 0.587]} & 0.704 \scriptsize{[0.607, 0.799]} & \textbf{0.558 \scriptsize{[0.504, 0.623]}} & 0.733 \scriptsize{[0.634, 0.822]} & 0.550 \scriptsize{[0.502, 0.606]} \\
        allergy & 0.662 \scriptsize{[0.625, 0.697]} & \textbf{0.542 \scriptsize{[0.523, 0.563]}} & \textbf{0.663 \scriptsize{[0.622, 0.706]}} & 0.539 \scriptsize{[0.517, 0.565]} & 0.653 \scriptsize{[0.612, 0.690]} & 0.524 \scriptsize{[0.507, 0.545]} & 0.656 \scriptsize{[0.617, 0.695]} & 0.534 \scriptsize{[0.514, 0.557]} \\
        asthma & 0.587 \scriptsize{[0.554, 0.622]} & \textbf{0.525 \scriptsize{[0.510, 0.540]}} & 0.596 \scriptsize{[0.554, 0.634]} & 0.515 \scriptsize{[0.499, 0.533]} & 0.605 \scriptsize{[0.565, 0.642]} & 0.520 \scriptsize{[0.503, 0.539]} & \textbf{0.608 \scriptsize{[0.569, 0.647]}} & 0.518 \scriptsize{[0.502, 0.536]} \\
        hearing loss & 0.750 \scriptsize{[0.705, 0.799]} & 0.615 \scriptsize{[0.578, 0.652]} & 0.753 \scriptsize{[0.691, 0.809]} & 0.617 \scriptsize{[0.571, 0.664]} & 0.748 \scriptsize{[0.688, 0.807]} & 0.610 \scriptsize{[0.566, 0.657]} & \textbf{0.763 \scriptsize{[0.705, 0.821]}} & \textbf{0.619 \scriptsize{[0.574, 0.668]}} \\
        vision loss & 0.629 \scriptsize{[0.544, 0.706]} & 0.549 \scriptsize{[0.510, 0.593]} & 0.626 \scriptsize{[0.528, 0.725]} & 0.557 \scriptsize{[0.512, 0.615]} & \textbf{0.695 \scriptsize{[0.607, 0.771]}} & 0.549 \scriptsize{[0.504, 0.603]} & 0.694 \scriptsize{[0.613, 0.770]} & \textbf{0.562 \scriptsize{[0.513, 0.616]}} \\
        urinary incontinence & \textbf{0.809 \scriptsize{[0.710, 0.892]}} & 0.652 \scriptsize{[0.585, 0.720]} & 0.774 \scriptsize{[0.672, 0.863]} & 0.597 \scriptsize{[0.518, 0.680]} & 0.765 \scriptsize{[0.645, 0.873]} & 0.627 \scriptsize{[0.546, 0.724]} & 0.784 \scriptsize{[0.663, 0.890]} & \textbf{0.655 \scriptsize{[0.569, 0.757]}} \\
        none & 0.658 \scriptsize{[0.630, 0.686]} & 0.540 \scriptsize{[0.524, 0.556]} & 0.682 \scriptsize{[0.651, 0.710]} & 0.539 \scriptsize{[0.522, 0.560]} & 0.711 \scriptsize{[0.683, 0.741]} & 0.541 \scriptsize{[0.524, 0.563]} & \textbf{0.722 \scriptsize{[0.694, 0.751]}} & \textbf{0.552 \scriptsize{[0.532, 0.574]}} \\
        \rowcolor[gray]{.95} \textbf{Mean (Health Conditions)} & 0.696 \scriptsize{$\pm$ 0.088} & 0.573 \scriptsize{$\pm$ 0.052} & 0.724 \scriptsize{$\pm$ 0.093} & 0.589 \scriptsize{$\pm$ 0.068} & 0.744 \scriptsize{$\pm$ 0.091} & 0.610 \scriptsize{$\pm$ 0.086} & \textbf{0.757 \scriptsize{$\pm$ 0.092}} & \textbf{0.618 \scriptsize{$\pm$ 0.084}} \\
        \midrule
        \rowcolor[gray]{.85} \textbf{Overall Mean} & 0.696 \scriptsize{$\pm$ 0.106} & 0.581 \scriptsize{$\pm$ 0.110} & 0.728 \scriptsize{$\pm$ 0.089} & 0.591 \scriptsize{$\pm$ 0.076} & 0.748 \scriptsize{$\pm$ 0.093} & 0.613 \scriptsize{$\pm$ 0.097} & \textbf{0.766 \scriptsize{$\pm$ 0.094}} & \textbf{0.626 \scriptsize{$\pm$ 0.098}} \\
        \bottomrule
    \end{tabular}}
\end{table}

\begin{figure*}[h!]
    \centering
    % Top Figure (AUROC)
    \includegraphics[width=0.76\linewidth, keepaspectratio]{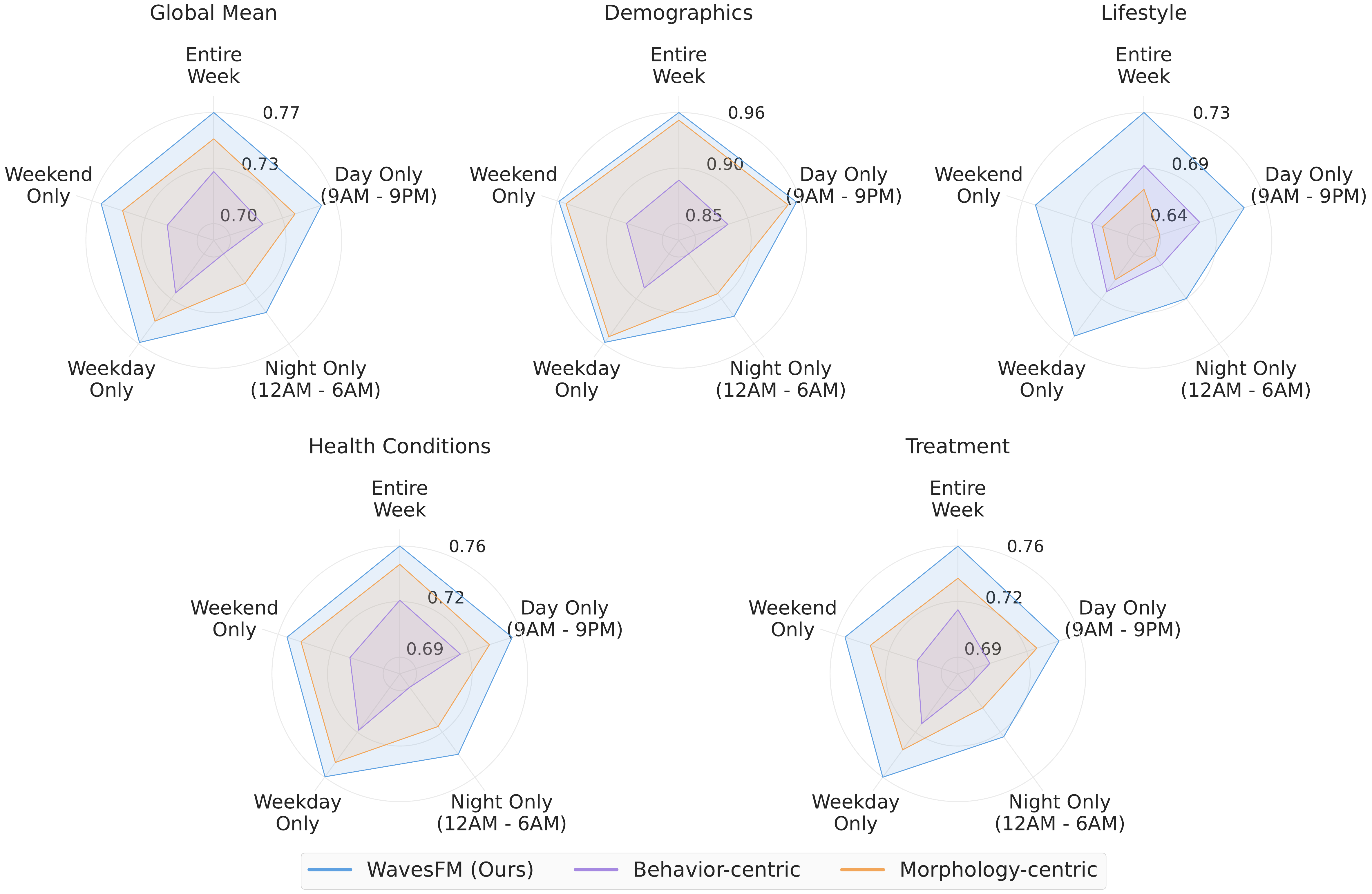}
    
    \vspace{0.5em} % Adjusts vertical spacing between the two plots
    
    % Bottom Figure (pAUROC)
    \includegraphics[width=0.76\linewidth, keepaspectratio]{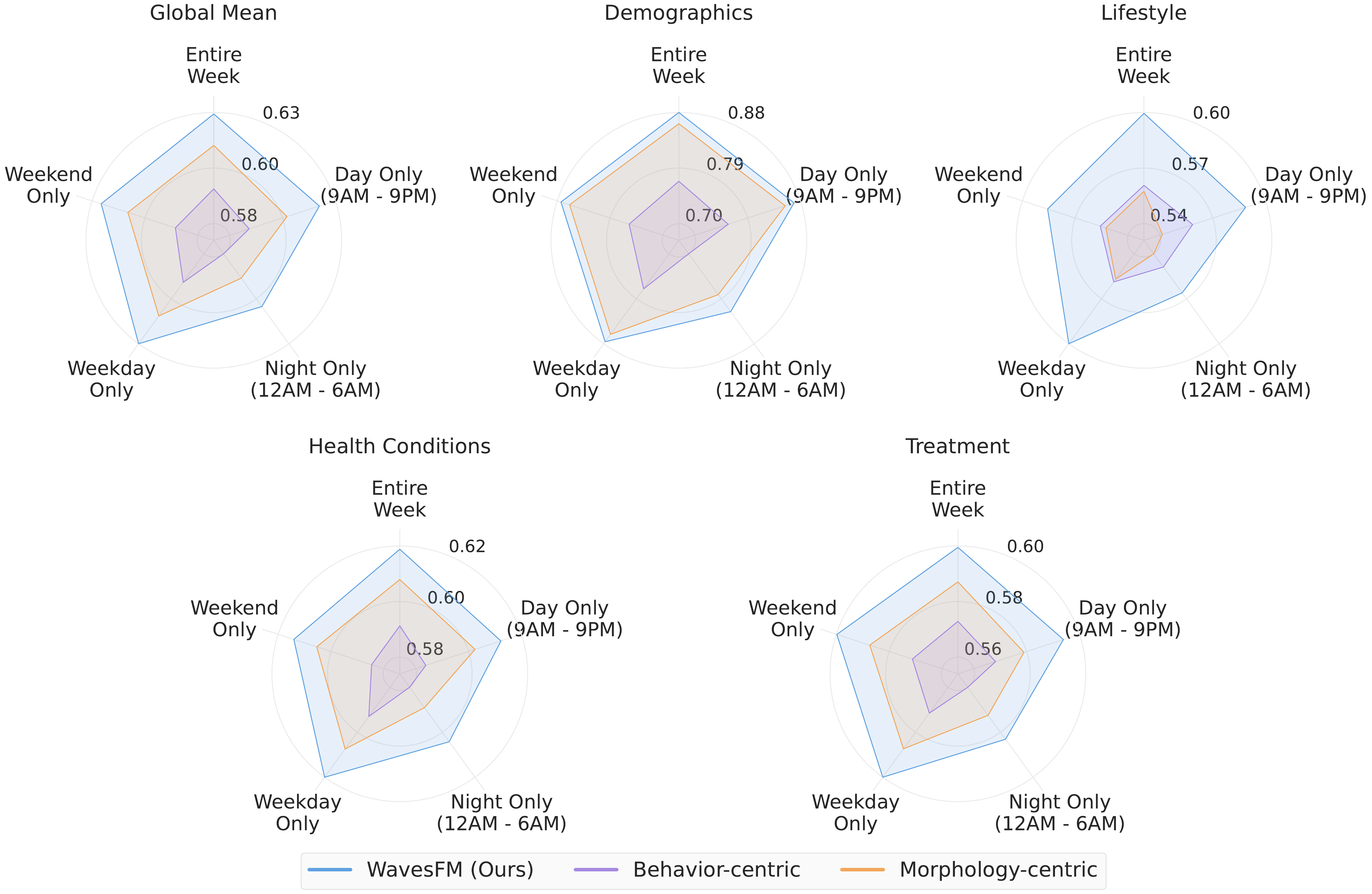}

    \caption{\textbf{Robustness to systematic missingness.}~Comparison of \texttt{WavesFM} against morphology- and behavior-centric baselines demonstrates its superior robustness across various missingness scenarios common in wearable data (e.g., weekday-only recording, night-only recording, etc.), evaluated using AUROC (top) and pAUROC (bottom).}
    \label{fig:ppgfm_radar_robustness}
    \vspace{-1em}
\end{figure*}

\subsection{Experiments on Longitudinal Missiness}
\label{sec:missingness}
In \Cref{fig:ppgfm_radar_robustness}, we visualize the complete AUC and pAUC performance metrics for the systematic longitudinal missingness experiments, detailing results across all four task categories alongside overall performance. Generally, both \texttt{WavesFM} and the morphology-centric baseline demonstrate fair robustness to missing data, except during nighttime periods where both models experience a performance drop. The behavior-centric baseline, in addition to yielding lower initial performance on the full dataset, exhibits a much more significant drop during nighttime and weekend periods, particularly on health condition (4.7\% drop vs 2.3\% for \texttt{WavesFM} at night) and treatment tasks (1.9\% drop vs 0.8\% for \texttt{WavesFM} on weekends). Conversely, an opposite trend emerges in lifestyle tasks, where the behavior-centric baseline shows greater robustness than the morphology-centric baseline when restricted to strictly daytime (1.8\% vs 6.4\% drop) or nighttime data (5.0\% vs 6.2\% drop), although both still underperform \texttt{WavesFM}.

\subsection{Comparison with Supervised Baselines}
\Cref{tab:supervised_baseline_detail} presents the classification performance of \texttt{WavesFM} using linear probing compared against three fully supervised baselines (see Appendix~\ref{subsec:supervised_learning_implementation} for implementation details on supervised baselines), with all models trained on a restricted 10\% subset of the labeled training data. Note that three tasks were excluded from this analysis due to an insufficient prevalence of positive labels in the reduced training set. Overall, \texttt{WavesFM} significantly outperforms the supervised baselines on over 80\% of the tasks (45 out of 55), achieving an average AUC improvement of 7.8\% over the closest baseline.
\begin{table}[p]
    \vspace{-4mm}
    \centering
    \caption{\textbf{Supervised Baseline vs. WavesFM Comparison Results.}}
    \label{tab:supervised_baseline_detail}
    \renewcommand{\arraystretch}{1.1}
    \scriptsize
    \resizebox{0.8\textwidth}{!}{%
    \begin{tabular}{ll cccccc|cc}
        \toprule
        \multirow{2}{*}{\textbf{Task}} & \multirow{2}{*}{\textbf{\# Pos}} & \multicolumn{2}{c}{\textbf{Cross Attention}} & \multicolumn{2}{c}{\textbf{LSTM}} & \multicolumn{2}{c}{\textbf{Weighted Pooling}} & \multicolumn{2}{c}{\textbf{\texttt{WavesFM}}} \\
        \cmidrule(lr){3-4} \cmidrule(lr){5-6} \cmidrule(lr){7-8} \cmidrule(lr){9-10}
        & & AUROC$^\uparrow$ & pAUC$^\uparrow$ & AUROC$^\uparrow$ & pAUC$^\uparrow$ & AUROC$^\uparrow$ & pAUC$^\uparrow$ & AUROC$^\uparrow$ & pAUC$^\uparrow$ \\
        \midrule
        \multicolumn{10}{l}{\textit{\textbf{Demographics}}} \\
        obesity (BMI $\geq$ 30) & 727 & 0.889 & 0.751 & 0.878 & 0.746 & 0.822 & 0.669 & \textbf{0.913} & \textbf{0.787} \\
        older age (age $\geq$ 50) & 447 & 0.932 & 0.805 & 0.929 & 0.785 & 0.919 & 0.760 & \textbf{0.951} & \textbf{0.829} \\
        sex male & 1254 & 0.991 & 0.965 & 0.988 & 0.957 & 0.952 & 0.859 & \textbf{0.992} & \textbf{0.966} \\
        \rowcolor[gray]{.95} \textbf{Mean (Demographics)} & --- & 0.937 & 0.840 & 0.932 & 0.829 & 0.898 & 0.763 & \textbf{0.952} & \textbf{0.860} \\
        \midrule
        \multicolumn{10}{l}{\textit{\textbf{Lifestyle}}} \\
        active smoker & 107 & 0.696 & 0.557 & 0.668 & 0.520 & 0.646 & 0.540 & \textbf{0.765} & \textbf{0.599} \\
        always salt & 94 & \textbf{0.606} & 0.505 & 0.555 & 0.500 & 0.580 & 0.509 & 0.593 & \textbf{0.520} \\
        freq alcohol & 351 & 0.672 & 0.550 & 0.604 & 0.525 & 0.615 & 0.517 & \textbf{0.781} & \textbf{0.635} \\
        freq sugar & 191 & 0.593 & 0.541 & 0.573 & 0.539 & 0.566 & 0.521 & \textbf{0.617} & \textbf{0.557} \\
        \rowcolor[gray]{.95} \textbf{Mean (Lifestyle)} & --- & 0.642 & 0.538 & 0.600 & 0.521 & 0.602 & 0.522 & \textbf{0.689} & \textbf{0.578} \\
        \midrule
        \multicolumn{10}{l}{\textit{\textbf{Treatment}}} \\
        ACE inhib. & 143 & 0.678 & 0.535 & 0.649 & 0.543 & 0.644 & 0.532 & \textbf{0.678} & \textbf{0.560} \\
        ARBs & 198 & 0.688 & 0.537 & 0.712 & 0.559 & 0.691 & 0.550 & \textbf{0.715} & \textbf{0.561} \\
        CCBs & 121 & 0.675 & 0.550 & 0.690 & 0.557 & 0.644 & 0.547 & \textbf{0.775} & \textbf{0.630} \\
        NSAIDs & 393 & \textbf{0.577} & 0.515 & 0.552 & 0.508 & 0.560 & \textbf{0.517} & 0.561 & 0.513 \\
        PDE inhib. & 91 & \textbf{0.655} & 0.513 & 0.634 & \textbf{0.517} & 0.577 & 0.513 & 0.655 & 0.516 \\
        aldosterone & 23 & 0.544 & 0.488 & 0.531 & \textbf{0.529} & 0.490 & 0.521 & \textbf{0.694} & 0.511 \\
        alpha blockers & 9 & 0.566 & 0.543 & 0.579 & 0.540 & 0.593 & 0.513 & \textbf{0.621} & \textbf{0.568} \\
        anti anxiety & 287 & 0.643 & 0.532 & 0.627 & 0.540 & 0.572 & 0.518 & \textbf{0.689} & \textbf{0.564} \\
        anti psychotics & 38 & 0.613 & 0.560 & 0.552 & 0.507 & 0.530 & 0.533 & \textbf{0.695} & \textbf{0.564} \\
        antcoagulants & 35 & 0.687 & 0.565 & 0.678 & 0.576 & \textbf{0.736} & 0.589 & 0.717 & \textbf{0.612} \\
        antidepressants & 306 & 0.672 & 0.558 & 0.675 & 0.566 & 0.681 & 0.566 & \textbf{0.767} & \textbf{0.621} \\
        antiplatelets & 17 & \textbf{0.735} & 0.539 & 0.651 & 0.519 & 0.694 & \textbf{0.601} & 0.661 & 0.584 \\
        beta blockers & 153 & 0.683 & 0.576 & 0.651 & 0.559 & 0.653 & 0.544 & \textbf{0.767} & \textbf{0.622} \\
        diuretics & 120 & 0.649 & 0.549 & 0.703 & 0.589 & 0.728 & 0.585 & \textbf{0.754} & \textbf{0.594} \\
        minoxidil & 75 & 0.602 & 0.498 & 0.590 & 0.507 & 0.606 & 0.529 & \textbf{0.635} & \textbf{0.544} \\
        none & 617 & 0.634 & 0.522 & 0.622 & 0.519 & 0.646 & \textbf{0.532} & \textbf{0.666} & 0.528 \\
        opioid & 27 & 0.662 & 0.545 & 0.567 & 0.568 & 0.642 & 0.528 & \textbf{0.784} & \textbf{0.642} \\
        sleeping aids & 227 & 0.560 & 0.522 & 0.571 & 0.532 & 0.585 & 0.526 & \textbf{0.603} & \textbf{0.536} \\
        \rowcolor[gray]{.95} \textbf{Mean (Treatment)} & --- & 0.640 & 0.536 & 0.624 & 0.541 & 0.626 & 0.541 & \textbf{0.691} & \textbf{0.571} \\
        \midrule
        \multicolumn{10}{l}{\textit{\textbf{Health Conditions}}} \\
        COPD & 31 & 0.568 & 0.489 & 0.614 & 0.495 & 0.592 & 0.519 & \textbf{0.662} & \textbf{0.543} \\
        allergy & 182 & 0.555 & 0.505 & 0.574 & 0.509 & 0.556 & 0.506 & \textbf{0.626} & \textbf{0.518} \\
        anxiety & 420 & 0.675 & 0.558 & 0.671 & 0.540 & 0.620 & 0.531 & \textbf{0.741} & \textbf{0.586} \\
        arthritis & 167 & 0.724 & 0.545 & 0.776 & 0.588 & 0.731 & 0.574 & \textbf{0.781} & \textbf{0.592} \\
        asthma & 223 & 0.529 & 0.504 & \textbf{0.567} & 0.507 & 0.551 & \textbf{0.515} & 0.552 & 0.511 \\
        atrial fibrillation & 42 & 0.755 & \textbf{0.647} & 0.757 & 0.629 & \textbf{0.763} & 0.622 & 0.757 & 0.622 \\
        back disorder & 145 & 0.639 & 0.539 & 0.648 & 0.540 & 0.613 & 0.524 & \textbf{0.681} & \textbf{0.563} \\
        cancer & 65 & 0.646 & 0.516 & 0.618 & 0.540 & 0.681 & \textbf{0.565} & \textbf{0.699} & 0.562 \\
        coronary artery disease & 32 & 0.762 & 0.618 & 0.772 & 0.649 & 0.801 & 0.651 & \textbf{0.850} & \textbf{0.687} \\
        depression & 422 & 0.669 & 0.556 & 0.657 & 0.553 & 0.648 & 0.534 & \textbf{0.724} & \textbf{0.590} \\
        diabetes & 125 & 0.687 & 0.556 & 0.716 & 0.576 & 0.737 & 0.581 & \textbf{0.791} & \textbf{0.620} \\
        hearing loss & 80 & 0.657 & 0.565 & 0.629 & 0.549 & 0.723 & 0.570 & \textbf{0.732} & \textbf{0.606} \\
        heart attack & 24 & 0.800 & \textbf{0.682} & 0.756 & 0.612 & 0.798 & 0.617 & \textbf{0.854} & 0.677 \\
        heart failure & 17 & 0.813 & 0.667 & 0.784 & 0.604 & 0.804 & 0.650 & \textbf{0.868} & \textbf{0.678} \\
 high cholesterol & 593 & 0.559 & 0.511 & 0.562 & \textbf{0.516} & 0.594 & 0.511 & \textbf{0.601} & 0.513 \\
        high triglycerides & 260 & 0.547 & 0.505 & 0.561 & 0.511 & 0.555 & 0.517 & \textbf{0.571} & \textbf{0.518} \\
        hypertension & 743 & 0.665 & 0.544 & 0.676 & 0.562 & 0.690 & 0.552 & \textbf{0.719} & \textbf{0.578} \\
        joint replace. & 32 & 0.766 & 0.614 & 0.727 & 0.600 & 0.790 & \textbf{0.636} & \textbf{0.847} & \textbf{0.636} \\
        kidney disease & 60 & 0.556 & \textbf{0.565} & 0.570 & 0.544 & 0.615 & 0.542 & \textbf{0.673} & 0.544 \\
        neck disorder & 65 & 0.616 & 0.528 & 0.513 & 0.519 & 0.558 & 0.518 & \textbf{0.664} & \textbf{0.574} \\
        neuropathy & 52 & 0.688 & 0.563 & 0.669 & 0.573 & 0.665 & 0.545 & \textbf{0.741} & \textbf{0.597} \\
        none & 348 & 0.605 & 0.533 & 0.639 & 0.535 & 0.660 & 0.540 & \textbf{0.678} & \textbf{0.547} \\
        osteoporosis & 24 & \textbf{0.781} & 0.594 & 0.759 & 0.619 & 0.729 & 0.625 & 0.779 & \textbf{0.656} \\
        other rhythm & 88 & 0.594 & 0.527 & 0.548 & 0.504 & 0.562 & 0.517 & \textbf{0.635} & \textbf{0.537} \\
        pacemaker & 4 & \textbf{0.916} & \textbf{0.737} & 0.894 & 0.731 & 0.812 & 0.560 & 0.890 & \textbf{0.737} \\
        stroke & 11 & 0.616 & 0.563 & 0.589 & \textbf{0.601} & 0.590 & 0.512 & \textbf{0.740} & 0.533 \\
        thyroid disease & 96 & 0.632 & 0.527 & 0.673 & 0.528 & 0.658 & 0.538 & \textbf{0.686} & \textbf{0.548} \\
        urinary incontinence & 20 & 0.700 & 0.517 & 0.769 & 0.614 & 0.743 & 0.596 & \textbf{0.809} & \textbf{0.648} \\
        valve disease & 12 & 0.504 & 0.474 & 0.547 & 0.505 & \textbf{0.700} & \textbf{0.595} & 0.514 & 0.526 \\
        vision loss & 40 & 0.538 & 0.512 & 0.508 & 0.507 & 0.463 & 0.498 & \textbf{0.649} & \textbf{0.540} \\
        \rowcolor[gray]{.95} \textbf{Mean (Health Condition)} & --- & 0.659 & 0.559 & 0.658 & 0.562 & 0.667 & 0.559 & \textbf{0.717} & \textbf{0.586} \\
        \midrule
        \rowcolor[gray]{.95} \textbf{Overall Mean} & --- & 0.667 & 0.565 & 0.658 & 0.567 & 0.661 & 0.561 & \textbf{0.719} & \textbf{0.595} \\
        \bottomrule
    \end{tabular}}
\end{table}

\subsection{Comparison with Public PPG Foundation Models}
\label{sec:public-ppg-fm}
\begin{figure}[h!]
	\centering
	\includegraphics[width=0.85\linewidth, keepaspectratio]{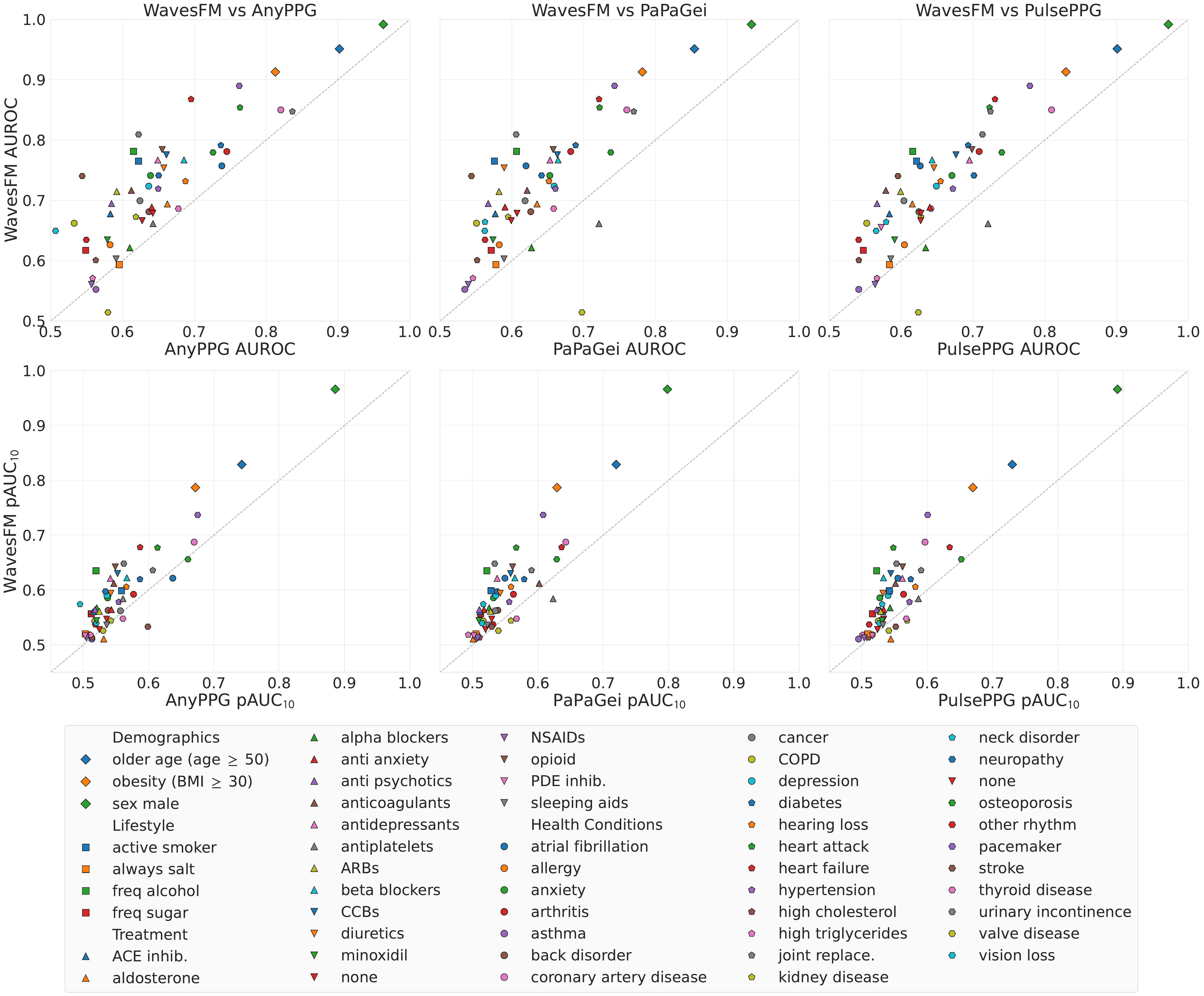}
	\caption{\textbf{Performance comparison with Foundation Model Baselines.}}
	\label{fig:ppgfm_auroc}
\end{figure}

In \Cref{fig:ppgfm_auroc}, we show detailed comparisons between \texttt{WavesFM} and three other public foundation models under an identical linear probing setup, restricting all models to 10\% downstream training split due to long inference time. The corresponding numerical results are provided in \Cref{tab:results_against_ppgfm}. With nearly all markers falling above the diagonal in these visualizations (49 out of 55 tasks), we can clearly observe \texttt{WavesFM}'s consistent advantage. The only notable exceptions are antiplatelets and valve disease, potentially due to the expected variance associated with small positive sample sizes in the restricted training set ($N=9$ and $N=12$, respectively).

\begin{table}[p]
    \centering
    \caption{\textbf{Detailed performance comparison against PPG Foundation Models.}}
    \label{tab:results_against_ppgfm}
    \scriptsize
    \renewcommand{\arraystretch}{1.1}
    \resizebox{0.8\textwidth}{!}{%
    \begin{tabular}{ll cccccc|cc}
        \toprule
        \multirow{2}{*}{\textbf{Task}} & \multirow{2}{*}{\textbf{\# Pos}} & \multicolumn{2}{c}{\textbf{\texttt{AnyPPG}} \cite{nie2025anyppg}} & \multicolumn{2}{c}{\textbf{\texttt{PulsePPG}} \cite{saha2025pulse}} & \multicolumn{2}{c}{\textbf{\texttt{PaPaGei}} \cite{pillai2025papagei}} & \multicolumn{2}{c}{\textbf{\texttt{WavesFM}}} \\
        \cmidrule(lr){3-4} \cmidrule(lr){5-6} \cmidrule(lr){7-8} \cmidrule(lr){9-10}
        & & AUROC$^\uparrow$ & pAUC$^\uparrow$ & AUROC$^\uparrow$ & pAUC$^\uparrow$ & AUROC$^\uparrow$ & pAUC$^\uparrow$ & AUROC$^\uparrow$ & pAUC$^\uparrow$ \\
        \midrule
        \multicolumn{10}{l}{\textit{\textbf{Demographics}}} \\
        obesity (BMI $\geq$ 30) & 727 & 0.813 & 0.672 & 0.829 & 0.670 & 0.782 & 0.629 & \textbf{0.913} & \textbf{0.787} \\
        older age (age $\geq$ 50) & 447 & 0.902 & 0.743 & 0.901 & 0.730 & 0.854 & 0.720 & \textbf{0.951} & \textbf{0.829} \\
        sex male & 1254 & 0.963 & 0.886 & 0.972 & 0.891 & 0.934 & 0.798 & \textbf{0.992} & \textbf{0.966} \\
        \rowcolor[gray]{.95} \textbf{Mean (Demographics)} & --- & 0.893 & 0.767 & 0.901 & 0.764 & 0.857 & 0.716 & \textbf{0.952} & \textbf{0.860} \\
        \midrule
        \multicolumn{10}{l}{\textit{\textbf{Lifestyle}}} \\
        active smoker & 107 & 0.622 & 0.559 & 0.622 & 0.542 & 0.576 & 0.528 & \textbf{0.765} & \textbf{0.599} \\
        always salt & 94 & \textbf{0.596} & 0.503 & 0.584 & 0.509 & 0.578 & 0.505 & 0.593 & \textbf{0.520} \\
        freq alcohol & 351 & 0.615 & 0.520 & 0.616 & 0.523 & 0.607 & 0.522 & \textbf{0.781} & \textbf{0.635} \\
        freq sugar & 191 & 0.549 & 0.512 & 0.547 & 0.516 & 0.571 & 0.512 & \textbf{0.617} & \textbf{0.557} \\
        \rowcolor[gray]{.95} \textbf{Mean (Lifestyle)} & --- & 0.595 & 0.523 & 0.592 & 0.522 & 0.583 & 0.517 & \textbf{0.689} & \textbf{0.578} \\
        \midrule
        \multicolumn{10}{l}{\textit{\textbf{Treatment}}} \\
        ACE inhib. & 143 & 0.583 & 0.515 & 0.584 & 0.532 & 0.577 & 0.511 & \textbf{0.678} & \textbf{0.560} \\
        ARBs & 198 & 0.592 & 0.525 & 0.599 & 0.529 & 0.582 & 0.527 & \textbf{0.715} & \textbf{0.561} \\
        CCBs & 121 & 0.661 & 0.552 & 0.676 & 0.544 & 0.663 & 0.558 & \textbf{0.775} & \textbf{0.630} \\
        NSAIDs & 393 & 0.557 & 0.505 & \textbf{0.564} & 0.504 & 0.539 & 0.508 & 0.561 & \textbf{0.513} \\
        PDE inhib. & 91 & 0.468 & 0.503 & 0.572 & 0.516 & 0.455 & 0.502 & \textbf{0.655} & \textbf{0.516} \\
        aldosterone & 23 & 0.663 & 0.531 & 0.616 & \textbf{0.544} & 0.635 & 0.501 & \textbf{0.694} & 0.511 \\
        alpha blockers & 9 & 0.610 & 0.521 & \textbf{0.634} & 0.543 & 0.627 & 0.525 & 0.621 & \textbf{0.568} \\
        anti anxiety & 287 & 0.641 & 0.543 & 0.640 & 0.533 & 0.591 & 0.517 & \textbf{0.689} & \textbf{0.564} \\
        anti psychotics & 38 & 0.585 & 0.518 & 0.567 & 0.523 & 0.567 & 0.510 & \textbf{0.695} & \textbf{0.564} \\
        antcoagulants & 35 & 0.612 & 0.547 & 0.579 & 0.551 & 0.621 & 0.602 & \textbf{0.717} & \textbf{0.612} \\
        antidepressants & 306 & 0.649 & 0.541 & 0.695 & 0.562 & 0.653 & 0.538 & \textbf{0.767} & \textbf{0.621} \\
        antiplatelets & 17 & 0.642 & 0.561 & 0.721 & 0.587 & \textbf{0.721} & \textbf{0.623} & 0.661 & 0.584 \\
        beta blockers & 153 & 0.686 & 0.567 & 0.643 & 0.533 & 0.664 & 0.565 & \textbf{0.767} & \textbf{0.622} \\
        diuretics & 120 & 0.658 & 0.542 & 0.646 & 0.532 & 0.589 & 0.542 & \textbf{0.754} & \textbf{0.594} \\
        minoxidil & 75 & 0.579 & 0.520 & 0.591 & 0.531 & 0.574 & 0.510 & \textbf{0.635} & \textbf{0.544} \\
        none & 617 & 0.627 & 0.525 & 0.627 & 0.524 & 0.599 & 0.520 & \textbf{0.666} & \textbf{0.528} \\
        opioid & 27 & 0.655 & 0.549 & 0.699 & 0.562 & 0.658 & 0.561 & \textbf{0.784} & \textbf{0.642} \\
        sleeping aids & 227 & 0.591 & 0.536 & 0.586 & 0.532 & 0.589 & 0.523 & \textbf{0.603} & \textbf{0.536} \\
        \rowcolor[gray]{.95} \textbf{Mean (Treatment)} & --- & 0.614 & 0.533 & 0.624 & 0.538 & 0.606 & 0.536 & \textbf{0.691} & \textbf{0.571} \\
        \midrule
        \multicolumn{10}{l}{\textit{\textbf{Health Conditions}}} \\
        COPD & 31 & 0.533 & 0.518 & 0.552 & 0.525 & 0.551 & 0.516 & \textbf{0.662} & \textbf{0.543} \\
        allergy & 182 & 0.583 & 0.508 & 0.605 & 0.516 & 0.583 & 0.507 & \textbf{0.626} & \textbf{0.518} \\
        anxiety & 420 & 0.639 & 0.537 & 0.671 & 0.527 & 0.653 & 0.532 & \textbf{0.741} & \textbf{0.586} \\
        arthritis & 167 & 0.745 & 0.577 & 0.709 & 0.564 & 0.682 & 0.562 & \textbf{0.781} & \textbf{0.592} \\
        asthma & 223 & \textbf{0.563} & \textbf{0.513} & 0.541 & 0.495 & 0.535 & 0.503 & 0.552 & 0.511 \\
        atrial fibrillation & 42 & 0.738 & \textbf{0.637} & 0.627 & 0.555 & 0.620 & 0.549 & \textbf{0.757} & 0.622 \\
        back disorder & 145 & 0.637 & 0.538 & 0.625 & 0.526 & 0.626 & 0.539 & \textbf{0.681} & \textbf{0.563} \\
        cancer & 65 & 0.624 & 0.557 & 0.604 & 0.526 & 0.619 & 0.535 & \textbf{0.699} & \textbf{0.562} \\
        coronary artery disease & 32 & 0.820 & 0.670 & 0.809 & 0.597 & 0.760 & 0.643 & \textbf{0.850} & \textbf{0.687} \\
        depression & 422 & 0.637 & 0.536 & 0.649 & 0.540 & 0.659 & 0.535 & \textbf{0.724} & \textbf{0.590} \\
        diabetes & 125 & 0.737 & 0.587 & 0.693 & 0.575 & 0.689 & 0.579 & \textbf{0.791} & \textbf{0.620} \\
        hearing loss & 80 & 0.688 & 0.566 & 0.655 & 0.582 & 0.652 & 0.559 & \textbf{0.732} & \textbf{0.606} \\
        heart attack & 24 & 0.764 & 0.614 & 0.723 & 0.548 & 0.722 & 0.567 & \textbf{0.854} & \textbf{0.677} \\
        heart failure & 17 & 0.695 & 0.587 & 0.731 & 0.634 & 0.721 & 0.636 & \textbf{0.868} & \textbf{0.678} \\
        high cholesterol & 593 & 0.563 & \textbf{0.513} & 0.541 & 0.510 & 0.552 & 0.511 & \textbf{0.601} & 0.513 \\
        high triglycerides & 260 & 0.558 & 0.511 & 0.566 & 0.501 & 0.546 & 0.493 & \textbf{0.571} & \textbf{0.518} \\
        hypertension & 743 & 0.650 & 0.554 & 0.672 & 0.573 & 0.661 & 0.556 & \textbf{0.719} & \textbf{0.578} \\
        joint replace. & 32 & 0.836 & 0.606 & 0.724 & 0.591 & 0.770 & 0.590 & \textbf{0.847} & \textbf{0.636} \\
        kidney disease & 60 & 0.619 & 0.542 & 0.628 & \textbf{0.569} & 0.595 & 0.559 & \textbf{0.673} & 0.544 \\
        neck disorder & 65 & 0.481 & 0.495 & 0.579 & 0.531 & 0.563 & 0.516 & \textbf{0.664} & \textbf{0.574} \\
        neuropathy & 52 & 0.650 & 0.534 & 0.701 & 0.543 & 0.641 & 0.537 & \textbf{0.741} & \textbf{0.597} \\
        none & 348 & 0.642 & 0.536 & 0.628 & 0.533 & 0.607 & 0.529 & \textbf{0.678} & \textbf{0.547} \\
        osteoporosis & 24 & 0.726 & \textbf{0.660} & 0.740 & 0.652 & 0.738 & 0.629 & \textbf{0.779} & 0.656 \\
        other rhythm & 88 & 0.550 & 0.519 & 0.541 & 0.511 & 0.563 & 0.531 & \textbf{0.635} & \textbf{0.537} \\
        pacemaker & 4 & 0.762 & 0.675 & 0.779 & 0.601 & 0.743 & 0.608 & \textbf{0.890} & \textbf{0.737} \\
        stroke & 11 & 0.544 & \textbf{0.599} & 0.596 & 0.552 & 0.544 & 0.529 & \textbf{0.740} & 0.533 \\
        thyroid disease & 96 & 0.678 & 0.561 & 0.642 & \textbf{0.568} & 0.658 & 0.567 & \textbf{0.686} & 0.548 \\
        urinary incontinence & 20 & 0.622 & 0.562 & 0.713 & 0.553 & 0.606 & 0.534 & \textbf{0.809} & \textbf{0.648} \\
        valve disease & 12 & 0.580 & 0.530 & 0.624 & \textbf{0.541} & \textbf{0.698} & 0.539 & 0.514 & 0.526 \\
        vision loss & 40 & 0.507 & 0.520 & 0.565 & 0.527 & 0.562 & 0.515 & \textbf{0.649} & \textbf{0.540} \\
        \rowcolor[gray]{.95} \textbf{Mean (Health Condition)} & --- & 0.646 & 0.562 & 0.648 & 0.552 & 0.637 & 0.550 & \textbf{0.717} & \textbf{0.586} \\
        \midrule
        \rowcolor[gray]{.95} \textbf{Overall Mean} & --- & 0.645 & 0.561 & 0.650 & 0.557 & 0.635 & 0.552 & \textbf{0.719} & \textbf{0.595} \\
        \bottomrule
    \end{tabular}}
\end{table}

\newpage

\begin{figure*}[h!]
    \centering

    \begin{subfigure}{0.32\linewidth}
        \centering
        \includegraphics[width=\linewidth, keepaspectratio]{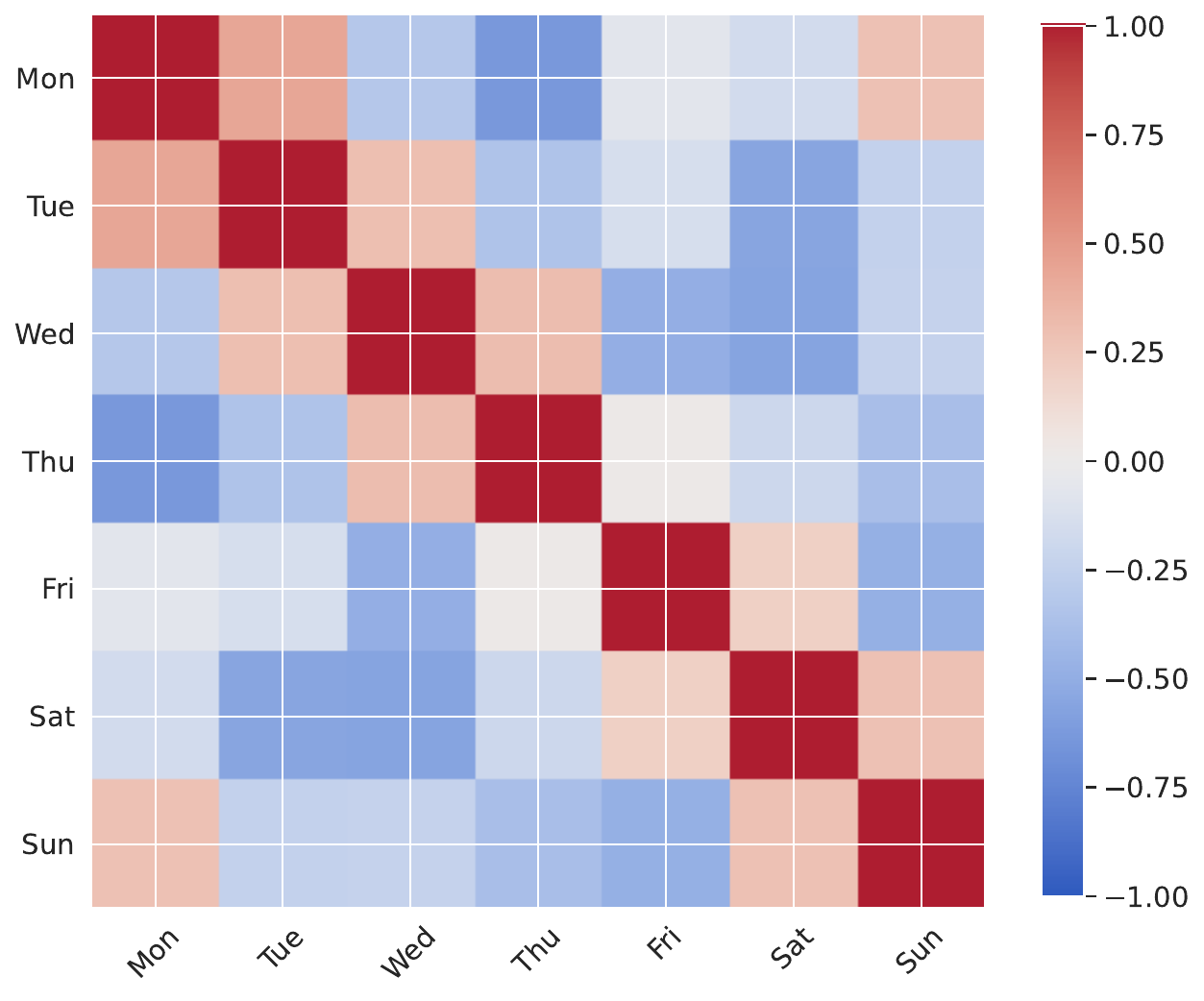}
        \caption{Day-of-week}
        \label{fig:positionalembedding_dayofweek}
    \end{subfigure}\hfill
    \begin{subfigure}{0.32\linewidth}
        \centering
        \includegraphics[width=\linewidth, keepaspectratio]{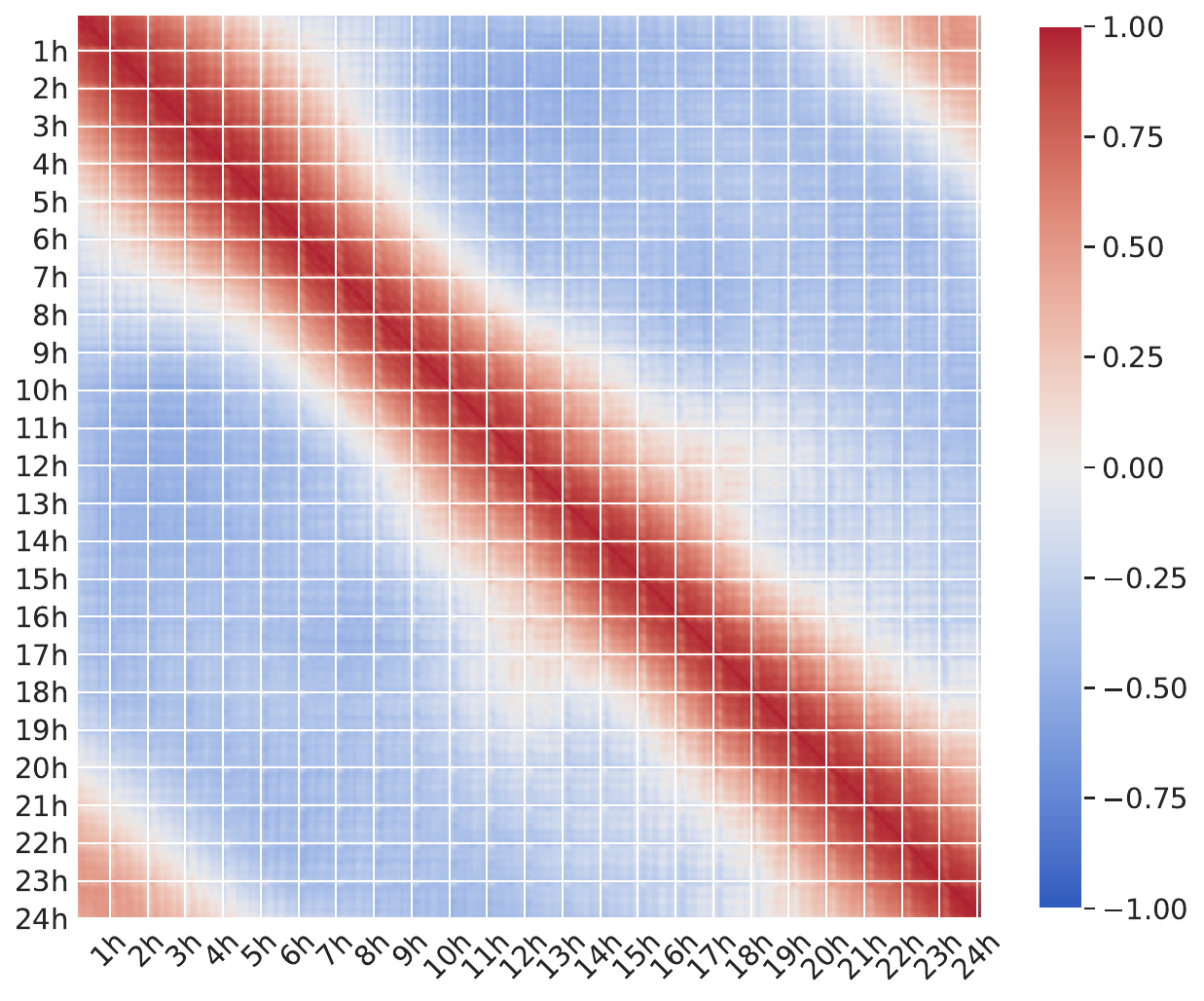}
        \caption{Time-of-day}
        \label{fig:positionalembedding_timeofday}
    \end{subfigure}\hfill
    \begin{subfigure}{0.32\linewidth}
        \centering
        \includegraphics[width=\linewidth, keepaspectratio]{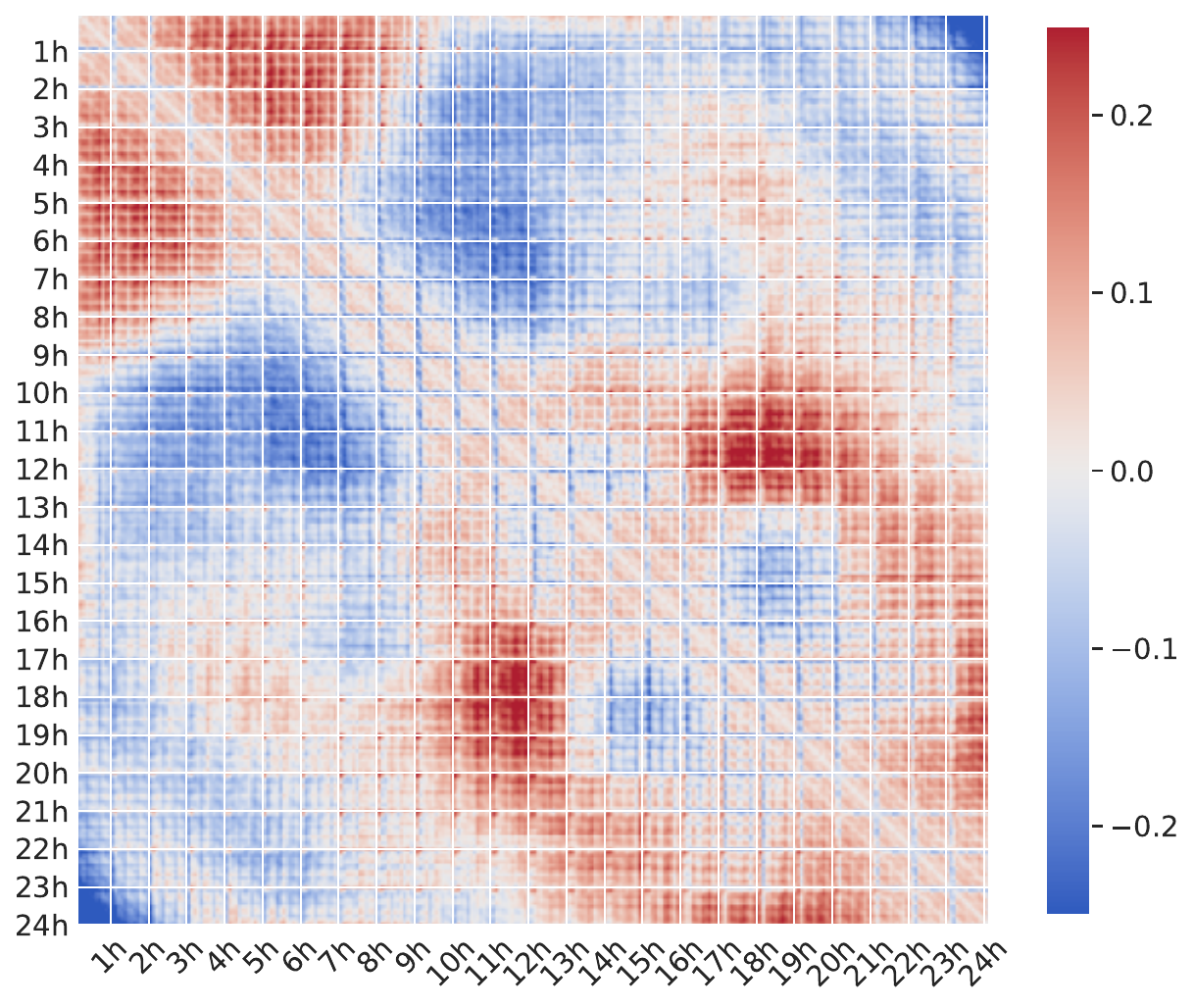}
        \caption{Time-of-day (Standardized)}
        \label{fig:positionalembedding_timeofday_standard}
    \end{subfigure}
    
    \caption{\textbf{Pairwise cosine similarity of factorized position embeddings.}}
    \label{fig:positionalembedding}
\end{figure*}

\begin{figure*}[h!]
    \centering
    \begin{subfigure}{0.32\linewidth}
        \centering
        \includegraphics[width=\linewidth, keepaspectratio]{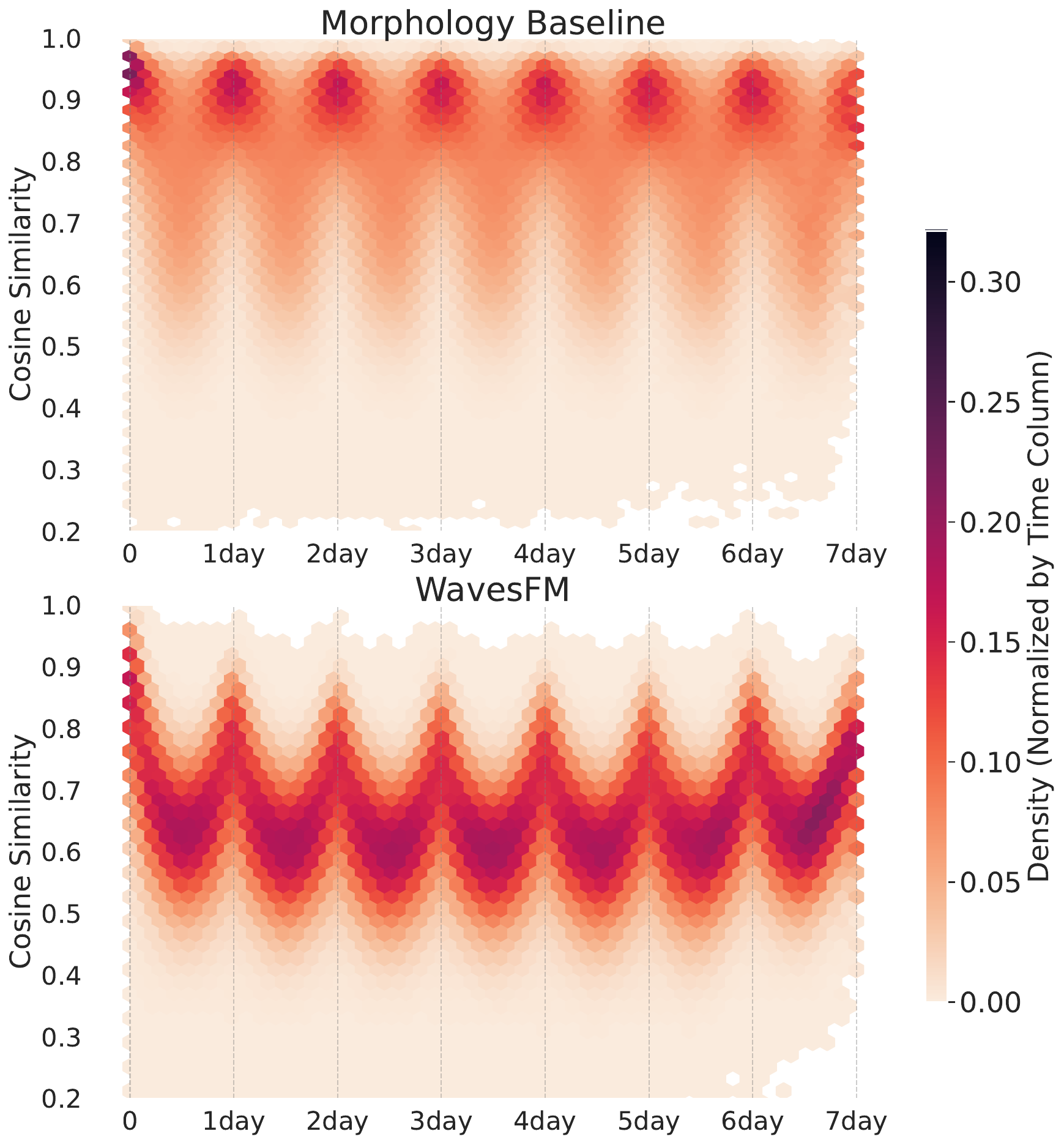}
        \caption{Pairwise similarity vs. time distance}
        \label{fig:circadian_noreference}
    \end{subfigure}\hfill
    \begin{subfigure}{0.32\linewidth}
        \centering
        \includegraphics[width=\linewidth, keepaspectratio]{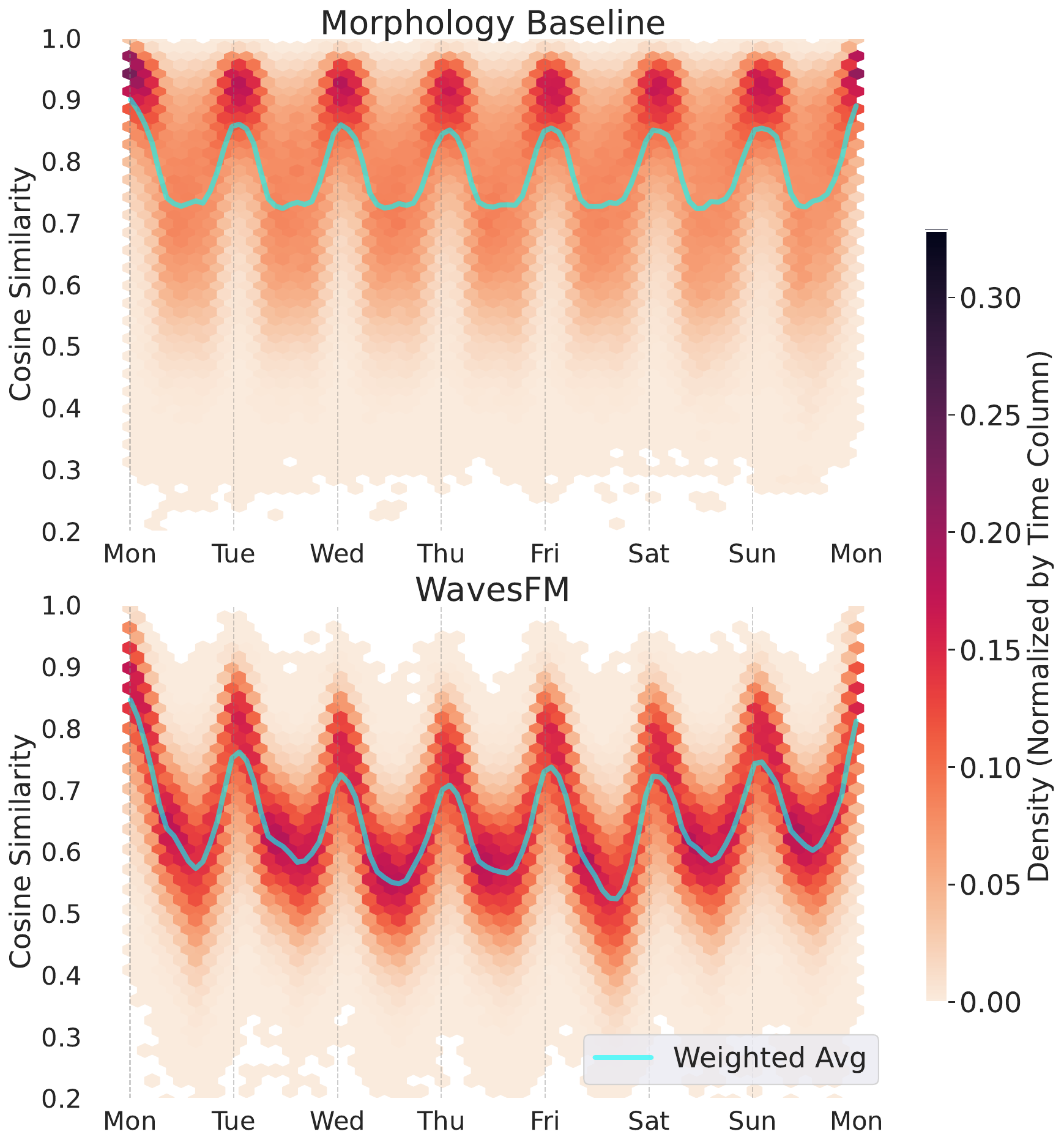}
        \caption{Similarity to Mon 0-1am reference}
        \label{fig:circadian_0am}
    \end{subfigure}\hfill
    \begin{subfigure}{0.32\linewidth}
        \centering
        \includegraphics[width=\linewidth, keepaspectratio]{figures/figure_main_consinesimilarity_9am.pdf}
        \caption{Similarity to Mon 9–10am reference}
        \label{fig:circadian_9am}
    \end{subfigure}
    
    \caption{\textbf{Supplementary analysis of circadian structure.} (a) Density of pairwise cosine similarities for all intra-subject segment pairs, plotted as a function of the time distance between the two segments. (b)(c) Density of pairwise cosine similarities across the week, calculated against each subject's Monday 0–1am  or 9-10am reference embeddings.}
    \label{fig:circadian_combined}
\end{figure*}
\subsection{Latent Space Analyses}
\label{sec:latent-space-analyses}
\textbf{Factorized Positional Embedding Analyses} We analyzed the factorized positional embeddings introduced in \Cref{subsec:method_stage_ii} by extracting the learned time-of-day ($288$ embeddings, 5-minute resolution) and day-of-week ($7$ embeddings, 1-day resolution) representations, computing their respective $288 \times 288$ and $7 \times 7$ pairwise cosine similarity matrices. The unnormalized time-of-day pairwise matrices (depicted in \Cref{fig:positionalembedding}) reveal strong temporal continuity, characterized by high similarity between adjacent temporal positions. However, the day-of-week embeddings exhibit a notable discontinuity between Thursday and Friday, potentially reflecting a shift in physiological rhythm at the end of the workweek.

To uncover latent variations obscured by the dominant diagonal continuity, we standardized the similarities by subtracting the mean similarity of pairs
with identical time gaps. This normalized view reveals distinct physiological regimes: a low-similarity cluster (blue) separates the daytime window (9am–2pm) from the night (12am–9am), indicating highly divergent states. Conversely, high-similarity clusters (red) emerge between discrete time windows that align with specific physiological states, such as sleep cycles (12am–3am vs. 4am–7am) and synchronized metabolic or dietary routines (10am–1pm vs. 5pm–8pm).

\textbf{Unpooled Segment Embedding Analyses}  Beyond positional embeddings, we directly analyze the similarities among unpooled segment embeddings to further evaluate the temporal encoder's contribution. Plotting pairwise cosine similarities against temporal distance (\Cref{fig:circadian_noreference}) reveals a strong circadian rhythm in \texttt{WavesFM} that is notably less pronounced in the morphology-centric baseline. Additionally, when substituting the main text's 9–10am reference with a 12–1am reference (\Cref{fig:circadian_0am}), the baseline yields the same shallow, binary pattern, where a shorter plateau likely reflects a shorter nocturnal sleep duration relative to daytime activity. In contrast, \texttt{WavesFM} consistently captures a continuous, full-cycle circadian rhythm independent of the chosen reference time.

\section{Ablation Experiments}
\label{sec:ablation}
\begin{figure*}[h!]
	\centering
	\includegraphics[width=0.9\linewidth, keepaspectratio]{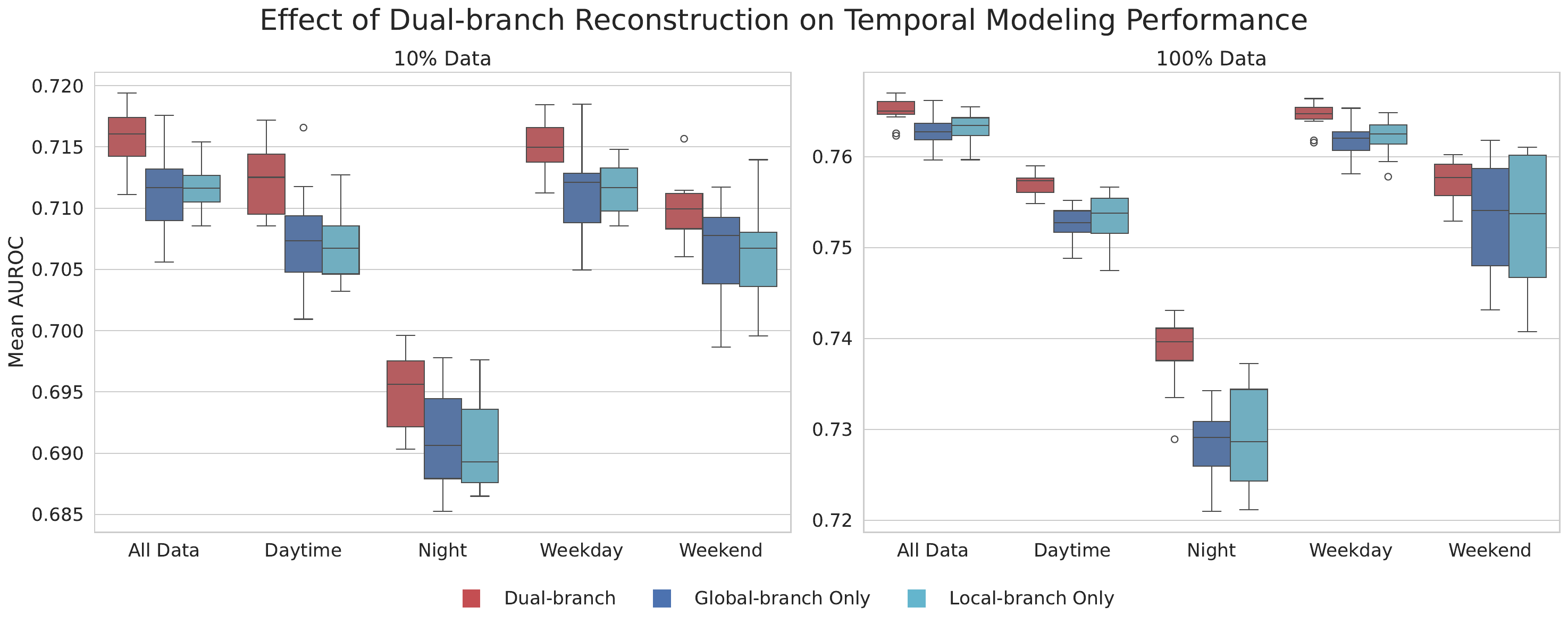}
	\caption{\footnotesize \textbf{Ablation of Dual-Branch Temporal Modeling}}
	\label{fig:ablation_dualbranch}
\end{figure*}

\textbf{Dual-Branch Ablation.} To evaluate the Stage II dual branch's contribution, we ablated the local and global branches independently, retraining each variant using identical hyperparameter sets to prevent checkpoint bias. By comparing the distribution of mean AUROCs across all hyperparameter configurations, we evaluated linear probing performance over five temporally restricted subsets (all, day, night, weekday, weekend) using 10\% and 100\% linear probing subsets. The dual-branch model consistently outperforms both single-branch baselines (\Cref{fig:ablation_dualbranch}). This advantage widens significantly under data constraints, whether restricted at the subject level (e.g., using only 10\% of subjects) or the segment level (e.g., night-only inputs), emphasizing that both branches are necessary for robust representations.

\begin{figure*}[h!]
    \centering
    \begin{minipage}[t]{0.4\linewidth} 
        \centering
        \includegraphics[width=\linewidth, keepaspectratio]{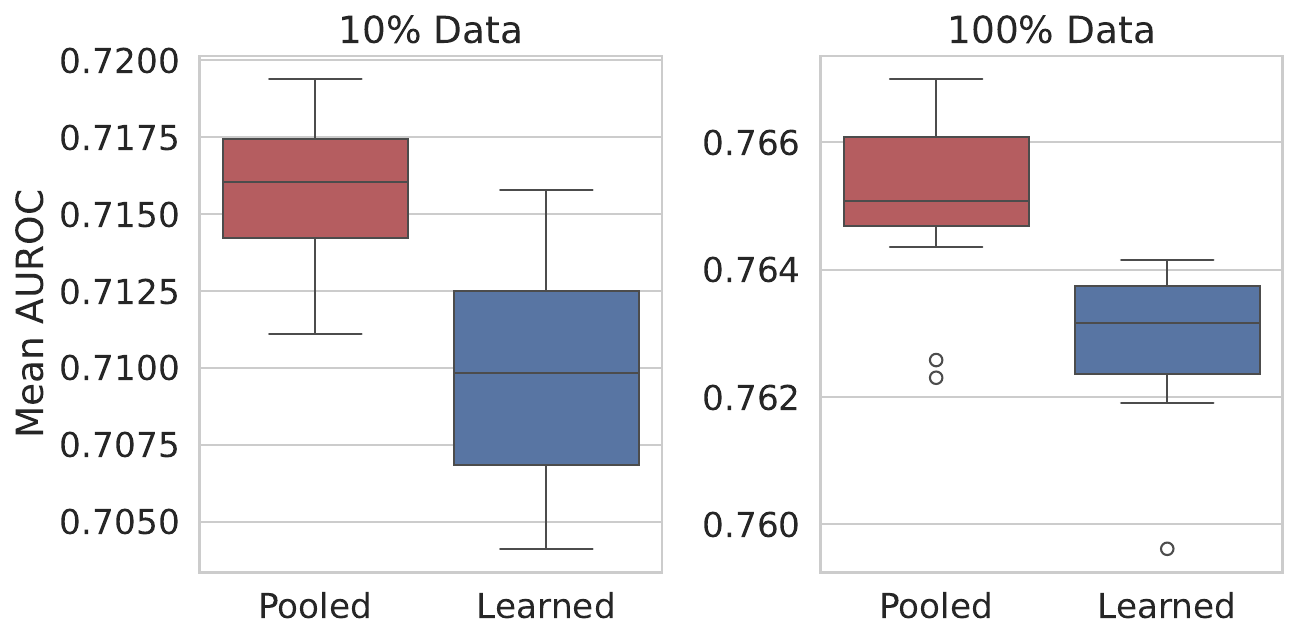}
        \caption{\footnotesize \textbf{Week-level Embedding Ablation}}
        \label{fig:ablation_cls}
    \end{minipage}
    \hspace{1cm}
    \begin{minipage}[t]{0.4\linewidth}
        \centering
        \includegraphics[width=\linewidth, keepaspectratio]{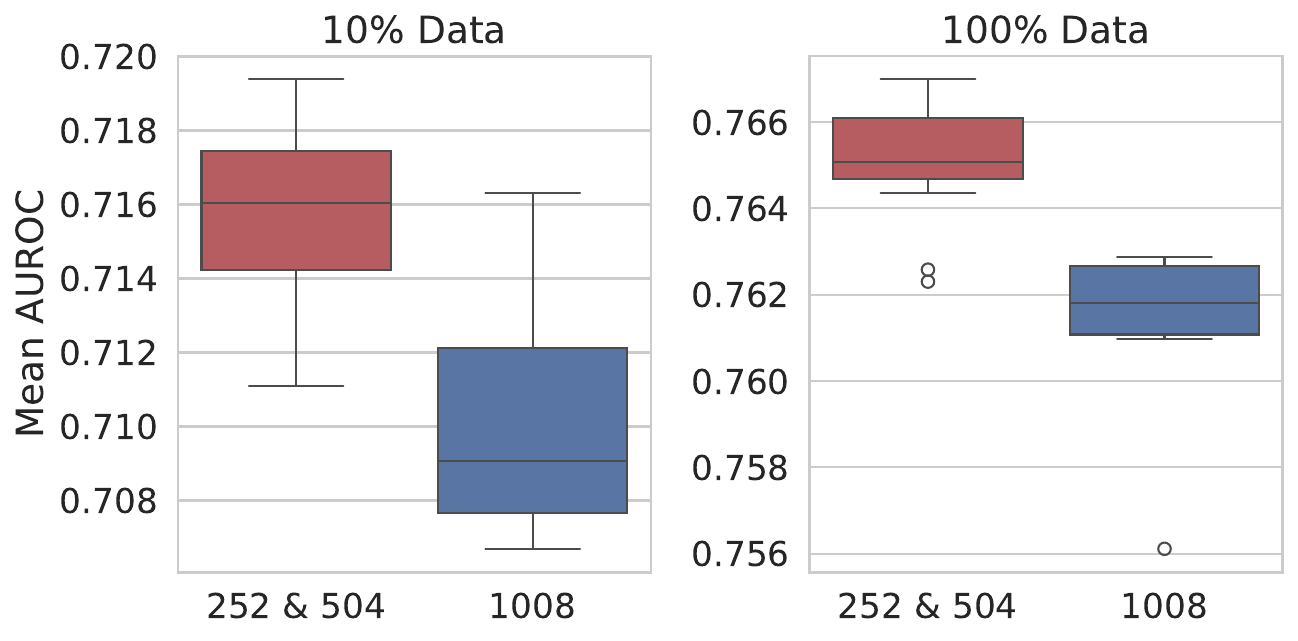}
        \caption{\footnotesize \textbf{Context Set Size Ablation}}
        \label{fig:ablation_context}
    \end{minipage}
\end{figure*}

\textbf{Week-level Embedding Ablation.} We evaluated an average-pooled token against a dedicated learnable token to serve as the week-level representation in dual-branch training. The pooled embedding strategy demonstrates a consistent performance advantage over the learnable token across both the 10\% and 100\% linear probing subsets (\Cref{fig:ablation_cls}).

\textbf{Context Set Size Ablation.} We trained the temporal encoder using three different context set sizes ($N_{\text{ctx}}=252$, $504$, and $1008$), corresponding to 12.5\%, 25\%, and 50\% of the total sequence length, respectively. We observe a significant performance advantage when using smaller context sizes (252 and 504) compared to the largest (1008) (\Cref{fig:ablation_context}). This aligns with existing literature, which suggests that higher masking ratios yield more robust representations in masked reconstruction pretraining~\cite{he2022masked}. These performance gains saturate at a context size of 504 (25\%) in our experiments, with no significant difference observed between the 252 and 504 configurations.

\end{document}